MIT AI Risk Initiative

FutureTech
THE ECONOMIC AND TECHNICAL FOUNDATIONS OF PROGRESS IN COMPUTING

Massachusetts Institute of Technology

# The AI Risk Repository

## A meta-review, database, and taxonomy of risks from artificial intelligence

**March 2026**

**Published at 10.1016/j.patter.2026.101517**

Peter Slattery, Alexander Saeri, Emily Grundy, Jess Graham, Michael Noetel, Risto Uuk, James Dao, Soroush Pour, Stephen Casper, and Neil Thompson

# The AI Risk Repository: A Meta-Review, Database, and Taxonomy of Risks From Artificial Intelligence

The living database is available at airisk.mit.edu.

Peter Slattery[1], Alexander K. Saeri[1,2], Emily A. C. Grundy[1,2], Jess Graham[1,2], Michael Noetel[1,2*], Risto Uuk[3,4], James Dao[5], Soroush Pour[5], Stephen Casper[6], and Neil Thompson[1].

[1]MIT FutureTech, Massachusetts Institute of Technology, Cambridge, MA 02139, USA

[2]School of Psychology, The University of Queensland, St Lucia, QLD 4072, Australia

[3]Future of Life Institute, Brussels 1040, Belgium

[4]KU Leuven, 3000 Leuven, Belgium

[5]Harmony Intelligence, Sydney, Australia

[6]Computer Science and Artificial Intelligence Laboratory, Massachusetts Institute of Technology, Cambridge, MA 02139, USA

Peter Slattery 0000-0002-6083-378X

Alexander K. Saeri 0000-0001-9254-0300

Emily A. C. Grundy 0000-0002-9633-1116

Jess Graham 0009-0003-9650-2792

Michael Noetel 0000-0002-6563-8203

Risto Uuk 0000-0002-2933-4608

James Dao 0000-0003-2596-9531

Soroush Pour 0009-0003-6156-5654

Stephen Casper 0000-0003-0084-1937

Neil Thompson 0000-0001-9888-573X

* Corresponding Author and Lead Contact: m.noetel@uq.edu.au; The University of Queensland, St Lucia, Queensland, 4072, Australia

# Summary

The risks posed by Artificial Intelligence concern academics, auditors, policymakers, AI companies, and the public. Researchers, policymakers, and technology companies discuss AI risks using inconsistent terminology—the same word may describe different problems, while different words describe identical concerns. This fragmentation impedes coordinated responses to AI challenges. We address this by creating the AI Risk Repository: a living database of 1,725 risks extracted from 74 existing taxonomies and frameworks. We organise these risks using two complementary classification systems. The Causal Taxonomy classifies risks by their origins: which entity causes them (human or AI), whether intentional, and when they occur (before or after deployment). The Domain Taxonomy classifies risks by their effects across seven areas, from discrimination and privacy violations to misinformation and weapons development. This shared reference enables more coordinated approaches to discussing, researching, auditing, and governing AI systems across sectors and jurisdictions.

# Introduction

For humanity to reap the benefits of AI, it must understand and address its risks. The range of possible risks is wide, from plagiarism [1] to pandemics [2]. Understanding and categorizing AI risks is fundamental to technological forecasting and anticipating the societal trajectory of artificial intelligence. As AI systems become increasingly autonomous and capable, the ability to forecast their potential impacts requires a comprehensive understanding of the risk landscape. Naturally, these risks have drawn considerable attention from academics, regulators, policymakers, and the public [3–6]. However, that broad attention has led to a diverse and disparate set of taxonomies, classifications, and other lists of AI risks. This paper aims to collate all of these taxonomies and harmonize them into one living resource to hold and organize those risks.

Having a range of partially overlapping risk frameworks can lead to confusion and hide vulnerabilities. For example, organizations developing AI models may present risk-mitigation plans that lack detail [7] or address relatively few risks [cf. 8–10]. Similarly, risk evaluators may be less able to comprehensively evaluate and report on AI risks without a clear understanding of the full range of threats [cf. 11]. A full taxonomy can help companies, governments, and model developers prioritize and know where to place controls.

Another challenge arising from multiple, overlapping frameworks is the conceptual ambiguity akin to psychology's 'jingle jangle fallacies' [12], where people use the same name for different risks, or different names for the same risk. For example, "privacy" might refer to a model's ability to leak sensitive information from the training data [13] or being free from government surveillance [14], which are very different risks. Our repository addresses this by providing a richer clustering of risk terms, mapping how different frameworks carve up the risk landscape and enabling practitioners to identify which specific risks underlie broader categories. We aim to provide additional structure that supports cross-framework comparison and comprehensive coverage. Shared understanding can reduce confusion and promote research usage, cross-study comparison, and the development of cumulative knowledge [e.g., 15,16]. Coherent frameworks are also important in legal, political, and practical settings, where they are often cited as goals for regulatory processes [17]. For example, the U.S.-EU Trade and Technology Council (TTC) stated in its joint roadmap for Trustworthy AI and Risk Management, "Shared terminologies and taxonomies are essential for operationalizing trustworthy AI and risk management in an interoperable fashion" [18].

Previous attempts to find shared understanding have usually drawn from narrative reviews, rather than the result of a systematic search. The few systematic reviews that exist [for exceptions: 19,20]have focused on specific categories of AI system (generative AI and Artificial General Intelligence [AGI], respectively) rather than risks from AI systems broadly. Existing taxonomies vary considerably in their adherence to best practice criteria for classification systems [21]. We observed that many taxonomies prioritise comprehensiveness over mutual exclusivity (with risks spanning multiple categories), while others achieve parsimony at the cost of exhaustive coverage. Few taxonomies describe explicit revision processes, and most are descriptive rather than explanatory in orientation. The number of competing taxonomies inadvertently makes it challenging to integrate relevant research into a cohesive shared understanding.

In this paper, we aim to address these limitations. We systematically reviewed existing AI risk classifications, frameworks, and taxonomies. We extracted the categories and subcategories of risks from included reports into a living database that we have updated over time. We applied a "best fit" framework synthesis approach [22,23] to develop two complementary taxonomies. The Causal Taxonomy captures antecedent conditions for risk: which entity's actions led to the risk, whether intentional, and when in the development lifecycle it occurs. The Domain Taxonomy captures consequent harms: the domains of impact such as loss of control, weapons development, privacy, economic harm, and so on. Together, these taxonomies allow risks to be classified by both their origins and their effects. We sought to rigorously extract AI risk frameworks into a comprehensive, extensible, and categorized risk database. It creates a foundation for a more coherent and complete approach to managing the risks posed by AI systems.

This work makes several theoretical contributions to technological forecasting and social change. First, we provide a comprehensive empirical foundation for theories of emerging technology risk by systematically analyzing 74 frameworks encompassing 1,725 distinct risks. This moves beyond single-framework approaches to create a meta-theoretical structure for understanding technological risk.

Second, our dual-taxonomy approach—combining causal and domain-based classifications—advances theoretical understanding of how technological risks emerge and manifest in society. The Causal Taxonomy provides a descriptive framework for categorising how existing taxonomies attribute risk sources. We understand this framework simplifies a more complex reality: many risks emerge from interactions between human decisions and AI system behaviours rather than from either in isolation [24]. Our 'Other' category (21% of coded risks) captures cases where this attribution was ambiguous or explicitly interactional, though existing taxonomies rarely theorised these interaction effects explicitly.

Third, we found that certain risk categories appear in relatively few frameworks (e.g., AI welfare appeared in 3% of frameworks; multi-agent risks in 7%). This suggests that future taxonomies might need to expand coverage as new risks like these are identified.

# Results

## Systematic literature search

We retrieved 17,288 unique articles from our searches and expert consultations. Of these records, we screened 7,945. We excluded 9,343 via our stopping criteria while using ASReview, which used machine learning to determine when further screening was unlikely to yield relevant content. We assessed the full text of 91 articles. A total of 43 articles and reports met the eligibility criteria: 21 from our search, 13 from forwards and backwards searching, and 9 from expert suggestions (see Figure 1). A database of included records and extracted data is available on the Open Science Framework [25].

## Ongoing expert consultation

In the period May 2024-March 2025, we received recommendations to consider 44 documents. Of those, 22 met eligibility criteria. Between March 2025 and December 2025, we received 78 document recommendations and included 9 that met criteria.

## Characteristics of included documents

We included 74 documents: 25 peer-reviewed articles, 26 preprints, 8 conference papers, 15 reports. We mainly identified recent literature, with all but five (93%) of the included documents published later than 2020.

The included documents varied in the type or scope of Artificial Intelligence they focused on. In most cases, the type of AI was not explicitly defined (n = 27). Large language model was the next most common (n = 13), followed by Generative AI (n = 10), General-purpose AI (n = 8), Artificial General Intelligence (n = 4) and Machine Learning (n = 3). Other terms included "AI and Machine Learning" (always described in the document as "AI/ML"), AI assistant, algorithmic systems, frontier AI, advanced AI, conversational agents and embodied AI.

Together, the included documents presented a total of 1,725 risk categories (e.g., "Privacy risks") or sub-categories of risk (e.g., "Compromising privacy by leaking sensitive information" [26]). Not all documents presented eligible risk categories in sufficient detail to allow us to code them with our taxonomies; two included documents were not coded as having any distinct risk categories or framework [e.g., 27,28] so were not extracted. Our supplementary materials also include a database of all risks and included documents and frameworks (Supplementary Tables S1 and S10).

**Figure 1. PRISMA Flow Diagram for Systematic Literature Search, Ongoing Expert Consultation, and Screening**

Flow diagram illustrating the identification, screening, eligibility assessment, and inclusion of documents for the AI Risk Repository. The left pathway shows the systematic literature search, which identified 17,288 unique records from database searching (n = 13,171) and forwards/backwards searching with initial expert consultation (n = 4,192). Title and abstract screening using ASReview software reduced this to 91 full-text documents, of which 43 met inclusion criteria. The right pathway shows ongoing expert consultation (May 2024–December 2025), which identified 122 additional documents through recommendations, with 31 meeting inclusion criteria after full-text review. Combined, 74 documents were included in the final analysis. Exclusion reasons at the full-text stage included focus on risk assessment processes rather than risk classification, focus on single risk categories, wrong document type, and documents that were not original AI risk frameworks.

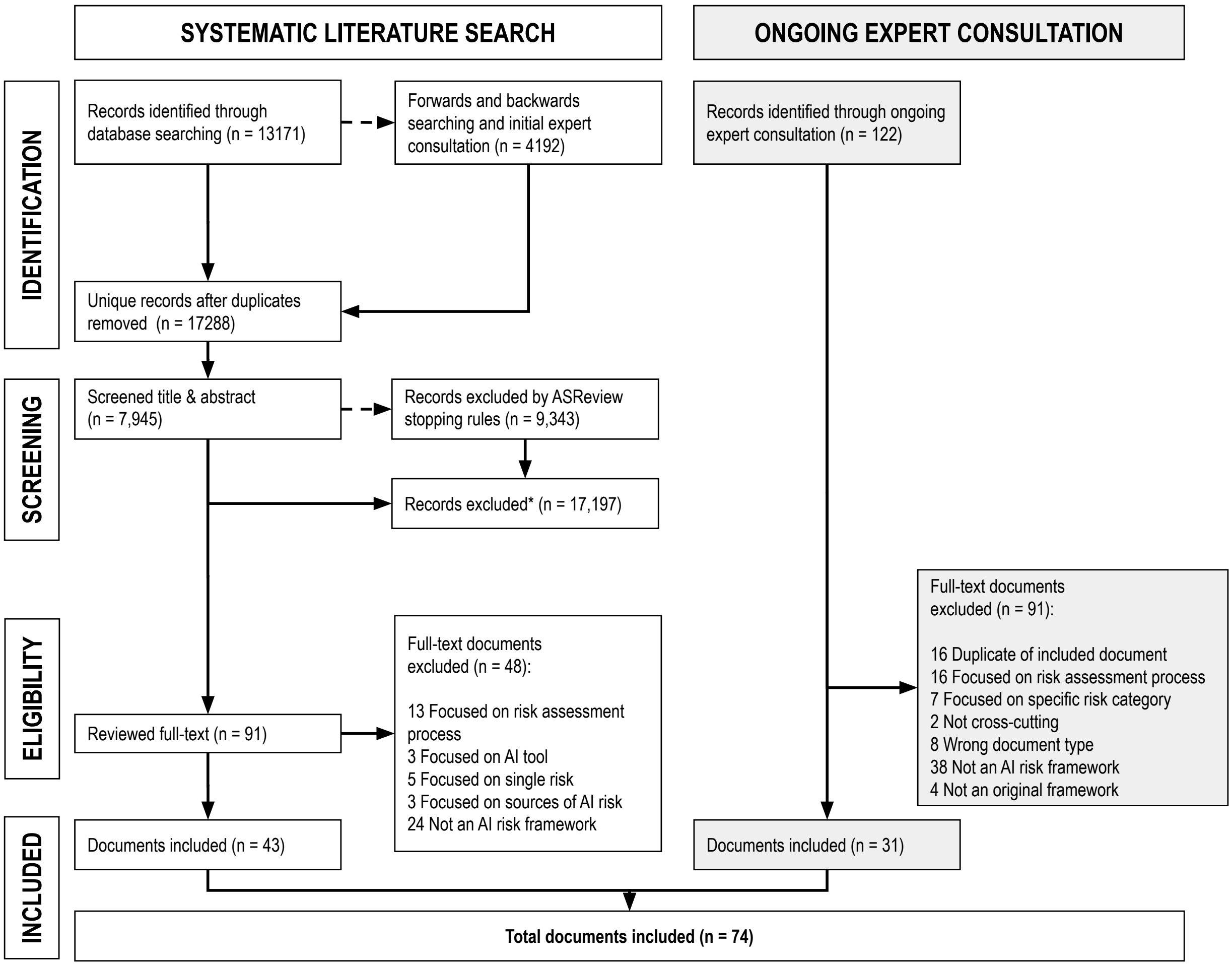

*Note. *The ASReview software was used to assist in screening titles and abstracts from searches.*

# Causal Taxonomy of AI Risks

We created our taxonomies using a best-fit framework synthesis. To create the Causal Taxonomy of AI Risks (Table 1), we selected and iterated on a taxonomy of causal factors [29]. This taxonomy uses Entity, Intent, and Timing to classify risks in the AI Risk Database. We coded 1,480 of the 1,725 (86%) potential risks extracted from our documents against the Causal Taxonomy. 136 did not present sufficient information to assess the Entity, Intent, or Timing, and 87 were discarded as they did not fit our definition of risk (e.g., where a previous taxonomy said “Governance - Regulation” without describing how the regulation itself was a risk [19]). Other reasons included the risk descriptions being too broad to code (e.g., ‘damage to political and economic institutions’). In supplementary materials, we provide detailed descriptions of each causal taxonomy variable (Supplemental Notes S1), report the iterative development of the best-fit taxonomy (Supplemental Notes S3), and coverage of the included documents across the Causal Taxonomy (Supplementary Tables S2, S3, and S4).

**Table 1. Causal Taxonomy of AI Risks**

The Causal Taxonomy classifies AI risks according to three categories of antecedent factors. Entity specifies whether the risk is caused by decisions or actions made by humans, by AI systems, or arises from human-AI interaction (or is ambiguously specified). Intent specifies whether the risk occurs as an expected outcome (intentional) or unexpected outcome (unintentional) of pursuing a goal, or is presented without clear specification of intentionality. Timing specifies whether the risk occurs before deployment (pre-deployment), after the AI model has been trained and deployed (post-deployment), or spans both phases (or is unspecified). Each risk is classified under exactly one level within each category. Of 1,725 extracted risks, 1,480 (86%) contained sufficient information to be coded against this taxonomy. See also Supplementary Table S1 for the full distribution of risks across taxonomy categories.

| Category | Level | Description |
|---|---|---|
| Entity | Human | The risk is caused by a decision or action made by humans |
| | AI | The risk is caused by a decision or action made by an AI system |
| | Other | The risk arises from human-AI interaction rather than either agent alone, or the causing entity is ambiguous or unspecified |
| Intent | Intentional | The risk occurs due to an expected outcome from pursuing a goal |
| | Unintentional | The risk occurs due to an unexpected outcome from pursuing a goal |
| | Other | The risk is presented as occurring without clearly specifying the intentionality |
| Timing | Pre-deployment | The risk occurs before the AI is deployed |
| | Post-deployment | The risk occurs after the AI model has been trained and deployed |
| | Other | The risk occurs across both pre- and post-deployment phases, or is presented without a clearly specified time of occurrence |

# Domain Taxonomy of AI Risks

To create the Domain Taxonomy of AI Risks, we used the best-fit framework synthesis to iterate on a taxonomy of categories of AI risk [26]. This taxonomy catalogs hazards and harms associated with AI. We coded 1,506 (87%) of the 1,725 potential risks extracted from our documents against the Domain Taxonomy. As above, the rest did not contain the necessary details or did not fit our definition of ‘risk’.

In Table 2, we present the Domain Taxonomy including subdomains and short descriptions of each subdomain. In supplementary materials, we present a detailed description of each subdomain using information from included documents (Supplemental Notes S2), as well as analysis of the distribution of risks and domains across the database (Supplementary Tables S5, S6, S7, S8, and S9). We also transparently report the evolution of the best-fit taxonomy.

**Table 2. Domain Taxonomy of AI Risks**

| Domain / Subdomain | Description |
|---|---|
| **1 *Discrimination & toxicity*** | |
| 1.1 Unfair discrimination and misrepresentation | Unequal treatment of individuals or groups by AI, often based on race, gender, or other sensitive characteristics, resulting in unfair outcomes and unfair representation of those groups. |
| 1.2 Exposure to toxic content | AI that exposes users to harmful, abusive, unsafe or inappropriate content. May involve providing advice or encouraging action. Examples of toxic content include hate speech, violence, extremism, illegal acts, or child sexual abuse material, as well as content that violates community norms such as profanity, inflammatory political speech, or pornography. |
| 1.3 Unequal performance across groups | Accuracy and effectiveness of AI decisions and actions is dependent on group membership, where decisions in AI system design and biased training data lead to unequal outcomes, reduced benefits, increased effort, and alienation of users. |
| **2 *Privacy & security*** | |
| 2.1 Compromise of privacy by obtaining, leaking, or correctly inferring sensitive information | AI systems that memorize and leak sensitive personal data or infer private information about individuals without their consent. Unexpected or unauthorized sharing of data and information can compromise user expectation of privacy, assist identity theft, or cause loss of confidential intellectual property. |
| 2.2 AI system security vulnerabilities and attacks | Vulnerabilities that can be exploited in AI systems, software development toolchains, and hardware, resulting in unauthorized access, data and privacy breaches, or system manipulation causing unsafe outputs or behavior. |
| **3 *Misinformation*** | |
| 3.1 False or misleading information | AI systems that inadvertently generate or spread incorrect or deceptive information, which can lead to inaccurate beliefs in users and undermine their autonomy. Humans that make decisions based on false beliefs can experience physical, emotional, or material harms |
| 3.2 Pollution of information ecosystem and loss of consensus reality | Highly personalized AI-generated misinformation that creates “filter bubbles” where individuals only see what matches their existing beliefs, undermining shared reality and weakening social cohesion and political processes. |
| **4 *Malicious actors & misuse*** | |
| 4.1 Disinformation, surveillance, and influence at scale | Using AI systems to conduct large-scale disinformation campaigns, malicious surveillance, or targeted and sophisticated automated censorship and propaganda, with the aim of manipulating political processes, public opinion, and behavior. |
| 4.2 Cyberattacks, weapon development or use, and mass harm | Using AI systems to develop cyber weapons (e.g., by coding cheaper, more effective malware), develop new or enhance existing weapons (e.g., Lethal Autonomous Weapons or chemical, biological, radiological, nuclear, and high-yield explosives), or use weapons to cause mass harm. |

| Domain / Subdomain | Description |
|---|---|
| 4.3 Fraud, scams, and targeted manipulation | Using AI systems to gain a personal advantage over others such as through cheating, fraud, scams, blackmail, or targeted manipulation of beliefs or behavior. Examples include AI-facilitated plagiarism for research or education, impersonating a trusted or fake individual for illegitimate financial benefit, or creating humiliating or sexual imagery. |

| Domain / Subdomain | Description |
|---|---|
| **5 *Human-computer interaction*** | |
| 5.1 Overreliance and unsafe use | Anthropomorphizing, trusting, or relying on AI systems by users, leading to emotional or material dependence and to inappropriate relationships with or expectations of AI systems. Trust can be exploited by malicious actors (e.g., to harvest information or enable manipulation), or result in harm from inappropriate use of AI in critical situations (e.g., medical emergency). Over reliance on AI systems can compromise autonomy and weaken social ties. |
| 5.2 Loss of human agency and autonomy | Delegating by humans of key decisions to AI systems, or AI systems that make decisions that diminish human control and autonomy, potentially leading to humans feeling disempowered, losing the ability to shape a fulfilling life trajectory, or becoming cognitively enfeebled. |
| **6 *Socioeconomic & environmental harm*** | |
| 6.1 Power centralization and unfair distribution of benefits | AI-driven concentration of power and resources within certain entities or groups, especially those with access to or ownership of powerful AI systems, leading to inequitable distribution of benefits and increased societal inequality. |
| 6.2 Increased inequality and decline in employment quality | Social and economic inequalities caused by widespread use of AI, such as by automating jobs, reducing the quality of employment, or producing exploitative dependencies between workers and their employers. |
| 6.3 Economic and cultural devaluation of human effort | AI systems capable of creating economic or cultural value, including through reproduction of human innovation or creativity (e.g., art, music, writing, coding, invention), destabilizing economic and social systems that rely on human effort. The ubiquity of AI-generated content may lead to reduced appreciation for human skills, disruption of creative and knowledge-based industries, and homogenization of cultural experiences. |
| 6.4 Competitive dynamics | Competition by AI developers or state-like actors in an AI “race” by rapidly developing, deploying, and applying AI systems to maximize strategic or economic advantage, increasing the risk they release unsafe and error-prone systems. |
| 6.5 Governance failure | Inadequate regulatory frameworks and oversight mechanisms that fail to keep pace with AI development, leading to ineffective governance and the inability to manage AI risks appropriately. |
| 6.6 Environmental harm | The development and operation of AI systems that cause environmental harm, such as through energy consumption of data centers or the materials and carbon footprints associated with AI hardware. |

| Domain / Subdomain | Description |
|---|---|
| **7 *AI system safety, failures & limitations*** | |
| 7.1 AI pursuing its own goals in conflict with human goals or values | AI systems that act in conflict with ethical standards or human goals or values, especially the goals of designers or users. These misaligned behaviors may be introduced by humans during design and development, such as through reward hacking and goal misgeneralisation, and may result in AI using dangerous capabilities such as manipulation, deception, or situational awareness to seek power, self-proliferate, or achieve other goals. |
| 7.2 AI possessing dangerous capabilities | AI systems that develop, access, or are provided with capabilities that increase their potential to cause mass harm through deception, weapons development and acquisition, persuasion and manipulation, political strategy, cyber-offense, AI development, situational awareness, and self-proliferation. These capabilities may cause mass harm due to malicious human actors, misaligned AI systems, or failure in the AI system. |
| 7.3 Lack of capability or robustness | AI systems that fail to perform reliably or effectively under varying conditions, exposing them to errors and failures that can have significant consequences, especially in critical applications or areas that require moral reasoning. |
| 7.4 Lack of transparency or interpretability | Challenges in understanding or explaining the decision-making processes of AI systems, which can lead to mistrust, difficulty in enforcing compliance standards or holding relevant actors accountable for harms, and the inability to identify and correct errors. |
| 7.5 AI welfare and rights | Ethical considerations regarding the treatment of potentially sentient AI entities, including discussions around their potential rights and welfare, particularly as AI systems become more advanced and autonomous. |
| 7.6 Multi-agent risks | Risks from multi-agent interactions due to incentives (which can lead to conflict or collusion) and/or the structure of multi-agent systems, which can create cascading failures, selection pressures, new security vulnerabilities, and a lack of shared information and trust. |

*The Domain Taxonomy classifies AI risks into seven domains and 24 subdomains based on the types of hazards and harms they describe. Domain 1 (Discrimination & toxicity) includes unfair discrimination, exposure to toxic content, and unequal performance across groups. Domain 2 (Privacy & security) covers privacy compromise and AI system security vulnerabilities. Domain 3 (Misinformation) addresses false information and pollution of the information ecosystem. Domain 4 (Malicious actors & misuse) encompasses disinformation at scale, cyberattacks and weapons, and fraud and manipulation. Domain 5 (Human-computer interaction) includes overreliance and loss of human agency. Domain 6 (Socioeconomic & environmental harm) covers power centralisation, inequality, devaluation of human effort, competitive dynamics, governance failure, and environmental harm. Domain 7 (AI system safety, failures & limitations) addresses misalignment, dangerous capabilities, lack of robustness, lack of transparency, AI welfare, and multi-agent risks. Unlike the Causal Taxonomy, domains are not mutually exclusive; some risks span multiple domains. Of 1,725 extracted risks, 1,506 (87%) were coded against this taxonomy. See also Supplementary Table S2 for subdomain descriptions with examples from included documents.*

## Overlap Between New and Existing Taxonomies

The Causal Taxonomy demonstrates that current frameworks cover multiple causal factors—the extracted risks were nearly equally attributed to AI systems (42%) versus human decisions (38%), and similarly distributed between intentional (35%) and unintentional (35%) causes. Frameworks tended to focus on post-deployment risks (62%), with fewer addressing pre-deployment risks (13%) and 25% either spanning both phases or not specifying timing.

The Domain Taxonomy analysis reveals substantial variation in the comprehensiveness of existing frameworks. While certain domains (e.g., AI System Safety) were represented in over 75% of the taxonomies we reviewed, categories such as AI welfare and rights (present in only 3% of frameworks) and Multi-agent risks (7% of frameworks) were rarely incorporated. Frameworks typically addressed only a subset of risks—averaging 8 of our identified 24 subdomains, with coverage ranging from 1 to 20 subdomains. Such variability indicates that existing taxonomies generally provide partial rather than comprehensive risk categorizations, potentially leaving significant gaps in risk assessment and management frameworks.

# Discussion

This paper presents attempts to rigorously curate AI risk frameworks into a comprehensive, extensible, and categorized risk database. We developed two taxonomies to classify 1,725 risks from 74 documents: the Causal Taxonomy of AI Risks (addressing how, when, and why risks emerge) and the Domain Taxonomy of AI Risks (categorizing risks into seven domains and 24 subdomains).

Our repository provides essential infrastructure for forecasting risks from emerging AI technologies that extend beyond current systems. The field is characterized by rapid growth—93% of included documents were published after 2020, with most appearing as preprints or conference papers. This suggests urgency in knowledge dissemination but poses challenges for coordination and standardization. Our extensible risk database aims to facilitate that coordination by continually arranging those risks into useful taxonomies.

As AI evolves toward greater autonomy, multimodal capabilities, and agentic behaviors, new risk categories will emerge at the intersections of our identified domains. For instance, the convergence of advanced language models with robotic systems may create novel risks spanning our 'AI system safety, failures & limitations' and 'Human-computer interaction' domains. Multi-agent risks appeared in only 7% of current frameworks. Given the trajectory toward AI systems that interact autonomously with each other and with human society,[30] future frameworks may benefit from greater attention to this category.

The Causal Taxonomy reveals patterns in where prior frameworks anticipate risks. The distribution of risks in our database shows greater attention to post-deployment risks (62%) than pre-deployment risks (13%). The appropriate distribution would depend on empirical evidence about risk frequency and severity that is beyond the scope of this review. Still, this distribution may warrant attention, because excessive focus on post-deployment risks may blind us to threats from development itself [31].

The Domain Taxonomy reveals uneven coverage across risk categories. While *AI System Safety, Failures & Limitations* and *Socioeconomic & Environmental Harms* are discussed in over 75% of documents, emerging areas like *AI welfare and rights* (3% of documents) and *Multi-agent risks* (7% of documents) are not yet integrated into frameworks, despite their potential significance as AI systems become more autonomous and interconnected. We hope one function of our review is to reduce the duplication (e.g., of new taxonomies including Socioeconomic & Environmental Harms) while aiming for comprehensive coverage (e.g., including emerging areas like AI welfare).

While our repository catalogs risks, it simultaneously enables innovation by providing clear boundaries within which beneficial AI development can proceed. Understanding the full risk landscape allows developers to channel innovation toward addressing genuine human needs while avoiding harmful pathways. For instance, awareness of discrimination risks has spurred innovation in fairness-aware machine learning, while understanding of AI misalignment drives research into frontier model evaluations and alignment. Our comprehensive mapping reveals 'safe harbors' for innovation—areas where risks are well-understood and manageable—as well as frontier domains requiring careful exploration. By making risks explicit and categorical, we reduce uncertainty for innovators and investors, potentially accelerating beneficial AI deployment. The repository thus serves a dual function: protecting society from AI harms while enabling the realization of AI's transformative potential for addressing societal challenges from healthcare to climate change.

# Practical Implications for Technology Management

Our AI Risk Repository provides concrete tools for managing AI's societal impact across multiple stakeholder groups:

- **For Technology Managers and AI Developers:** the repository supports comprehensive risk assessment during AI development and deployment. Organizations can use our causal analysis to identify opportunities for mitigations, recognizing that many "AI risks" actually require human intervention during design, development, or governance phases. The Domain Taxonomy enables systematic risk assessment during product development cycles. Organizations can use our 24 subdomains as a checklist during design reviews, ensuring comprehensive risk consideration before deployment. We recommend integrating our taxonomies into existing risk management frameworks, using them to structure safety cases and inform resource allocation decisions. For example, the Causal Taxonomy reveals that 13% of identified risks relate to the pre-deployment phase. Organisations concerned with development-phase safety may find this subset of the repository particularly relevant for structuring their risk assessments.
- **For Policymakers and Regulators:** the repository forms the basis for operationalizing vague regulatory references to "harm" and "risk." It can support compliance frameworks required by regulations like the EU AI Act [32] and facilitate international collaboration through shared terminologies—essential for initiatives like the EU-US Trade and Technology Council's efforts to develop interoperable AI governance [18]. The repository operationalizes vague regulatory concepts like "high-risk AI systems" by providing 1,725 specific risk examples organized into actionable categories. Regulators can use the Causal Taxonomy to design targeted interventions—for instance, focusing on the 37% of risks attributed to human decisions through training requirements, while addressing the 42% from AI system actions through technical standards.

- **For Auditors and Compliance Officers:** the repository addresses a critical gap: the absence of comprehensive frameworks for determining when and where AI systems pose specific risks. Current industry risk-management frameworks often address narrow risk subsets, making comprehensive evaluation challenging [33]. Our repository provides the foundation for developing objective standards necessary for comprehensive AI audits. The Domain Taxonomy's seven domains can structure audit protocols, while the repository's specific risk examples can inform testing scenarios. Organizations can demonstrate due diligence by documenting how they address each relevant subdomain.
- **For Researchers**: the taxonomies enable systematic synthesis across disparate studies and identification of knowledge gaps. The database can guide research prioritization—for instance, our finding that pre-deployment human-caused risks receive minimal attention despite emerging concerns about dangerous AI development. It may provide a framework for prioritising among the full range of AI risks, and help researchers identify the controls necessary to cover that range.

# Limitations and Future Directions

Several limitations warrant consideration. While we attempted to extract risks verbatim and conducted extensive calibration, conducting extraction and coding in duplicate would reduce the risk of bias or error. Our search, while comprehensive, focused on cross-cutting frameworks and excluded domain-specific taxonomies. Users focused on a particularly narrow domain (e.g., medical diagnostics) might benefit from using those focused taxonomies. We have not conducted a formal validation study to assess whether independent users can reliably categorise novel risks using our taxonomies. Future work should assess inter-rater reliability with coders external to the authorship team. Such validation would increase confidence in the taxonomies' utility for standardising risk discourse across organisations. Our Causal Taxonomy assigns risks to human or AI sources, which simplifies the sociotechnical reality that many risks emerge from human-AI interactions rather than from either agent in isolation [24]. Similarly, our Timing categories (pre-deployment, post-deployment, other) simplify a more complex reality. Modern AI development often involves continuous iteration: models are deployed, monitored, retrained, and updated in ongoing cycles rather than progressing through discrete phases.The repository's structure also trades some precision for comprehensiveness; while we aimed to capture all risks, we could not capture risk likelihood, severity, or interactions between risks.

The field would benefit from assessing the severity and likelihood of these risks, as has been done for catastrophic AI risks [34]. Doing so would help in prioritising among these risks, prioritising among proposed controls, and addressing underexplored areas. Future research will need to help actors understand which of these risks are most important for them to focus on. Once prioritized, it is still unclear how companies and governments should best mitigate those risks without stifling innovation. Future studies should aim to identify the best mitigations for the most important risks. Finally, any new frameworks should consider including a broader range of risks identified here. For example, most frameworks focus on language models rather than emerging concerns like agentic AI and multi-agent systems [14,20]. The limited attention to AI welfare and rights (appearing in only two documents) deserves attention until we can confidently rule out AI sentience for increasingly advanced systems [35].

## Conclusions

Our AI Risk Repository provides critical infrastructure for effective technology governance in an era of rapid AI advancement. The development of appropriate governance frameworks requires a shared understanding of what risks need to be governed—our repository provides this foundation. The repository enables adaptive governance approaches essential for managing rapidly evolving technologies. Rather than static regulatory frameworks, our living database supports dynamic policy-making that can respond to emerging risks while avoiding premature or overly broad restrictions.

The taxonomies bridge multiple research domains essential for understanding AI's societal impact. From a forecasting perspective, our systematic categorization of 1,725 risks provides the empirical foundation for projecting AI's future trajectories. From a management perspective, our taxonomies offer practical tools for organizational decision-making about AI development and deployment. From an impact assessment perspective, our comprehensive risk mapping enables evaluation of AI's multifaceted effects across social, economic, and environmental dimensions. From a governance perspective, our repository provides the shared terminology and categorization schemes necessary for effective policy development. This integration across domains is essential because AI risks themselves transcend traditional boundaries—a single AI system may simultaneously raise concerns about discrimination (social impact), competitive dynamics (economic impact), and existential risks.

The AI Risk Repository establishes a foundation for more coordinated approaches to understanding and managing AI risks. By making the database living and extensible, we enable continuous refinement as new risks emerge and understanding evolves. It provides a common reference point that can reduce conceptual confusion, guide research and policy priorities, and support the development of comprehensive governance frameworks. As AI capabilities advance rapidly, such shared understanding becomes essential for managing the risks so we can safely realize the benefits.

# Methods

A summary of our methodology is available in Figure 2. We used a systematic search strategy, forwards and backwards searching, and expert consultation to identify AI risk classifications, frameworks, and taxonomies. Since conducting the original systematic literature search, we have periodically identified additional relevant research through an ongoing expert consultation. We extracted the individual risks from these documents into a living AI Risk Database. We conducted two best-fit framework syntheses to create a Causal Taxonomy of AI Risks (see Table 1) and Domain Taxonomy of AI Risks (see Table 2), by adapting existing frameworks [26,29]. We did this by testing their effectiveness at coding our risk data and modifying them until we created a final version that could effectively code all relevant risks.

### Figure 2. Overview of Study Methodology

Schematic overview of the study methodology showing the progression from systematic literature search to taxonomy development. The systematic literature search comprised three components: peer-reviewed and grey literature search, forwards/backwards search and initial expert consultation, and ongoing expert consultation. These searches identified existing AI risk frameworks, which were extracted into a living AI risk database. Best-fit framework synthesis was then applied to develop two complementary outputs: the Causal Taxonomy (classifying risks by entity, intent, and timing) and the Domain Taxonomy (classifying risks by type of harm across seven domains).

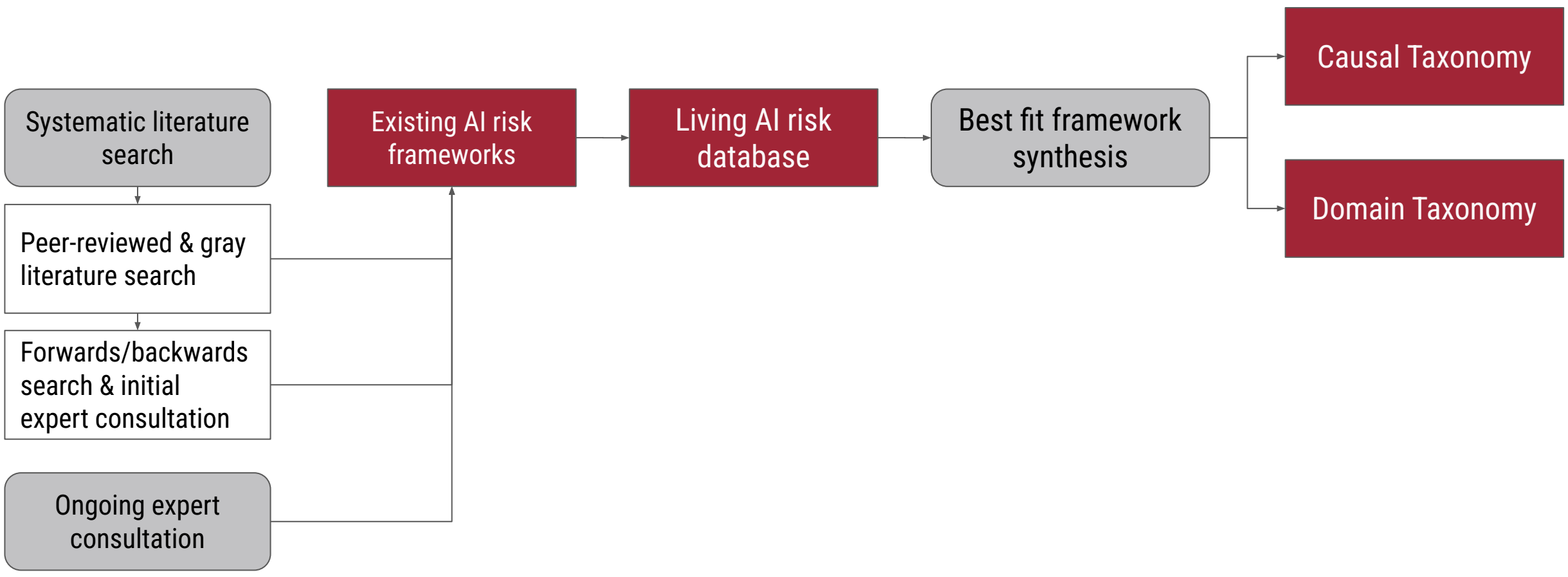

# Systematic literature search

We conducted this study as a systematic review [36,37]. The protocol was registered in advance using the Open Science Framework in April 2024 (https://osf.io/egzc8). We included reviews, articles, reports, and documents primarily focused on proposing new frameworks, taxonomies, or other structured classifications of risks from AI present across multiple locations and industry sectors. We followed the Society for Risk Analysis [38] in defining "AI risk" as "the possibility of an unfortunate occurrence associated with the development or deployment of artificial intelligence," while recognizing that this term can be defined in many ways [39,40].

We excluded book chapters, theses, commentaries, editorials, and protocols. Our pilot searches suggested that including these documents would significantly increase the number of search results to screen but not the number of relevant results. We excluded documents which discussed impacts, outcomes, or other consequences of AI without specifying specific risks because we were interested in risk classification.

Due to our interest in broad, structured classifications of risks from AI, we excluded documents which focused only on risks from AI that are present in a single location or sector or that discussed risks specific to particular risk categories (e.g., content solely focused on different types of unfair decision-making) or very specific AI tools (e.g., content solely focused on risks from DALL-E). We excluded content which merely cited or discussed existing theories, frameworks, models, taxonomies, and other structured classifications rather than proposing and explaining them. This was because we wanted to understand and extract specific risks using their original source material. We excluded anything which discussed sources of risk at a high level of abstraction (e.g., the sources of sociotechnical risk in AIs) or risk-assessment processes (e.g, how organizations can assess risks from AI) rather than focusing on classifying AI risks more specifically. Non-English articles, reports, and documents were excluded due to resource constraints related to their retrieval and translation.

Two of the above exclusion criteria were added after protocol registration in order to retain only the most relevant documents: (1) Focus only on one category of risk from AI; (2) Focus on sources of risk or the risk-assessment process.

## Search strategy

Our search strategy comprised two stages. In Stage 1 we conducted a systematic search of peer-reviewed and gray literature (i.e., non peer-reviewed materials) to identify relevant articles. We begin by explaining our search-term generation and strategy, followed by our database searches in Scopus and various preprint databases. We then describe our screening process, which uses active learning with ASReview. This process includes four phases: initial random screening for training data, application of active learning with specific stopping rules, model switching for comprehensive coverage, and quality evaluation. Finally, we outline our full-text screening and calibration procedures. In Stage 2 we conducted forwards (citation) and backwards (references) searching and expert consultation to identify additional eligible articles (both processes described below).

### Stage 1: Searching & screening peer reviewed and gray literature

#### Searching

Search terms were generated through an iterative process and chosen for their empirical balance between sensitivity and specificity [41]. This included terms related to artificial intelligence (Artificial intelligence, AI, Artificial general intelligence, AGI), frameworks, taxonomies, and other structured classifications (Framework, Review, Overview, Taxonomy*), and risks (Risk, Harm, Hazard). This led to the following search string:

> TITLE-ABS-KEY ( ( "artificial intelligence" OR ai OR "artificial general intelligence" OR agi ) AND ( framework OR taxonom* OR review ) AND ( risk OR harm OR hazard ) ) AND ( LIMIT-TO ( LANGUAGE, "English" ) ).

We conducted a Scopus search to identify relevant academic research. The same search string was used on the following preprint databases to identify relevant literature: arXiv, Social Science Research Network (SSRN), Research Square, medRxiv, TechRxiv, bioRxiv, and ChemRxiv. Both searches were conducted on 4 April 2024. Relevant articles were downloaded for screening.

**Title/abstract and full-text screening**

Two authors formed a team to conduct title/abstract and full-text screening. Before screening, the team calibrated their decision-making by independently screening the same randomly selected articles (n = 23), comparing the results, and resolving disagreements. Agreement was achieved on 21 of 23 records (91%). To expedite title and abstract screening, we used active learning in ASReview [42], with the process done independently and in duplicate by two reviewers.

Active learning is an emerging research technique which uses machine learning to reduce the total number of records requiring manual screening. It is now widely used for efficiently screening large datasets in systematic reviews and meta-analyses [43,44], and has been validated in a number of diverse fields [42,44] and datasets [45].

Throughout the active-learning process, we followed the four step SAFE procedure outlined by Boetje and van de Schoot [46] to ensure that screening identified relevant articles both rigorously and efficiently:

*Phase 1: Screen a random set of articles to create training data for active-learning model*

As per SAFE, the screening team each randomly screened and labeled 1% of the total search yield (264 records in total). Each member of the team then created separate projects in ASReview and uploaded their own files, which included all retrieved studies and the random screening data. The random screening data was automatically marked as prior knowledge, and the active-learning phase commenced.

*Phase 2: Apply active learning during screening until stopping rule is reached*

For the first iteration of the active-learning model, the team followed Boetje and van de Schoot's [46] recommendation to use the Oracle model and the default model setup (TF-IDF as the feature extractor, Naive Bayes as the classifier, maximum as the query strategy and dynamic resampling (double) as the balance strategy). We aimed to follow Boetje and van de Schoot's [46] four-fold stopping heuristics, screening until four mutually independent conditions are met:

1. All key papers are marked as relevant.
2. At least twice the estimated number of relevant records in the total dataset are screened.
3. More than 10% of the total dataset has been screened.
4. No relevant records are identified in the last 50 records screened.

These four stopping heuristics aim to achieve a sensitivity of 95% [44], ensuring comprehensive data assessment while preventing excessive time spent on unlikely candidates.

The team met three of these conditions: they i) screened more than twice the estimated number of relevant records, ii) screened more than 10% of the total dataset, and iii) had not identified any relevant records in the last 50 records. However, one condition ('All key papers are marked as relevant') could not be met due to a bug with the model. Only three out of the four key papers [20,26,47,48] had appeared in the screening process, and the final key paper was scheduled to appear several thousand papers later. Because Stage 3 of the SAFE process aims to ensure that records are not missed due to the initial model, the screening team switched models to see if a new model would locate the relevant paper.

*Phase 3: Switch active-learning model and screen additional records until stopping rule is reached*

Based on a review of relevant literature [42,e.g., 44], we use the Oracle model with the following set-up: a fully connected neural network (2 hidden layers) model as the classifier and sBert as the feature extractor, maximum as the query strategy and dynamic resampling (double) as the balance strategy. The model was trained on the data that was labeled while using the previous model. Screening stopped when no extra relevant records were identified in the last 50 records. Both authors screened in the missing key paper within the first two records found by the new model.

*Phase 4: Evaluate quality*

For quality checks, the screening team screened records previously labeled as irrelevant using the Oracle model and the default model set-up (i.e., the same model that was used in the initial/main model phase). This model was trained using the 10 highest- and lowest-ranked records from the model switching phase. Both team members screened records to identify any relevant records that might have been falsely excluded. This continued until the stopping rule was met (no extra relevant records identified in the last 50 records).

One member of the screening team screened the full text of all records that were included at the title/abstracts step. For calibration, 10% of the records were screened in duplicate, with 100% interrater reliability achieved. Conflicts were resolved by discussion for any remaining records.

### Stage 2: Forwards and backwards searching and expert consultation

Following full-text screening, we undertook forwards and backwards searching using Scopus, Google Scholar, and various preprint servers hosting the included gray literature. Backwards searching involved identifying and reviewing all references from articles included in Stage 1, while forwards searching involved identifying and reviewing all articles that cited an included article. We also undertook an expert consultation, which involved sharing the preliminary set of included articles with their authors and other experts and requesting recommendations for relevant frameworks that had been overlooked. All records identified during forwards and backwards searching and expert consultation were screened by one author. Those which met inclusion criteria were added to the backlog for extraction.

# Extraction into living AI Risk Database

Five authors were involved in data extraction. A template data extraction spreadsheet was developed to capture various details from the studies, including title, abstract, author, year, source/outlet, risk category name, risk category description, risk subcategory name, risk subcategory description, and page number. This spreadsheet was refined over several rounds of pilot testing and extractor calibration on subsets of randomly selected articles. Data extraction was then conducted individually, with regular meetings for discussion and conflict resolution. Based on the recommendations of grounded theory, we aimed to capture the studied phenomena directly from the data rather than impose our interpretations [49,cf. 50]. Consequently, we extracted risks based on how the authors presented them, maintaining fidelity to their original categorizations and descriptions.

# Best fit framework synthesis approach

Seven authors were involved in data synthesis. We used a “best fit” framework synthesis approach to develop two AI risk taxonomies. Best-fit framework synthesis is a method for rapidly, clearly, and practically understanding the relationships and structures between concepts in a topic area [22,23]. It combines the strengths of framework synthesis [51], which is a “top down” positivist method where concepts are coded against a pre-existing structure, and thematic synthesis [52], which is a “bottom up” interpretative method where concepts are iteratively analyzed to identify patterns and structure. We describe the process in Figure 3.

## Figure 3. Methodology for Best Fit Framework Synthesis

Flowchart illustrating the iterative best-fit framework synthesis process used to develop the Causal and Domain Taxonomies. The process began with included AI risk classifications, frameworks, and taxonomies. Two parallel initial steps involved defining the purpose of each taxonomy and extracting risks into the living AI risks database. An existing taxonomy was selected as the starting framework (v1). Risks from the database were coded using this existing taxonomy. Categories and descriptions were then updated to accommodate risks that could not be coded against v1, producing an updated taxonomy (v2). This coding and updating cycle was repeated iteratively (v3, etc.) until all relevant risks could be coded using the final taxonomy. See Supplementary Information for detailed documentation of taxonomy iterations.

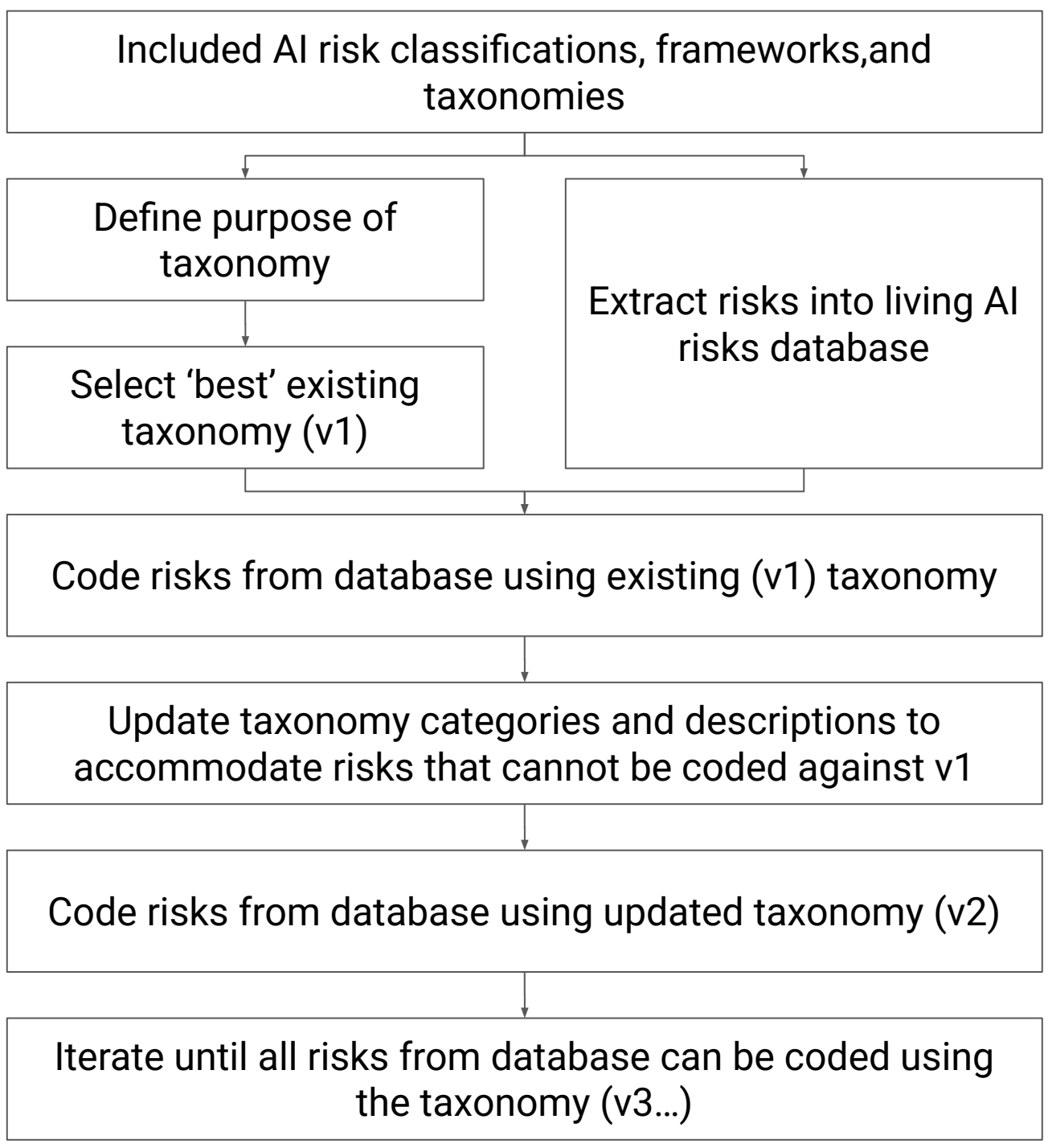

To conduct a best-fit framework synthesis, we identified published frameworks in an area (through our systematic search and screening), selected the “best” existing framework for our purpose, then used that existing framework to code the concepts (i.e., all the risks extracted into the living AI Risk

Database). Some risks could not be coded against the existing framework. We then conducted a secondary thematic analysis to identify new themes in those risks and determined what changes needed to be made to the framework to accommodate those themes. This involved updating the existing categories, creating new categories, or changing the structure of the framework. This process was repeated until achieving a final version of the framework that could most effectively code all relevant risks.

By starting with an existing framework, the synthesis can achieve a coherent framework more quickly than inductively or thematically analyzing all the individual concepts (risks) across all included papers. The trade-off is that the existing framework creates a particular "lens" for understanding and categorizing the individual concepts, which may lead to a disconnect between the synthesized findings and the theoretical or epistemological perspectives in the original and highly varied papers. In order to mitigate this, we attempted to code the risks based on the exact wording the authors had presented rather than our interpretation of what they may have intended to communicate [49,cf. 50].

## Why we developed two taxonomies of AI risk

The goal of our best-fit framework synthesis was to create a common frame of reference for understanding and addressing the risks from artificial intelligence. We found that authors implicitly or explicitly used different lenses [53,54,e.g., 55] to create their frameworks. These lenses reveal and obscure different aspects of the AI risk landscape.

Through our systematic search, we identified two types of frameworks, which we will refer to here as causal and domain frameworks. "Causal frameworks" focused on antecedents, capturing broad factors that specify how, when, or why an AI risk might emerge [e.g., 48,56] rather than discuss categories of specific hazards and harms. In contrast, "domain frameworks" focused on outcomes: specific hazards and harms [57,e.g., 58] but didn't explore their causes.

The differences here made it challenging to create a single framework. Often, specific domain risks did not fit into the categories within a causal framework, and the broad categories in those frameworks were insufficiently specified to be useful, in isolation, for creating shared understanding. Similarly, broad causal categorizations of how, when, or why an AI risk emerges did not fit with domain frameworks outlining narrow and specific sets of risk.

We therefore resolved that the ideal common frame of reference required two intersecting taxonomies: one to precisely decompose or define an AI risk based on the antecedent conditions under which it occurred (a "causal taxonomy"), and one that classified commonly discussed hazards and harms associated with AI into understandable and distinct domains (a "domain taxonomy").

In the following sections, we describe the process of developing these two taxonomies using a best-fit framework synthesis approach.

## Development of Causal Taxonomy of AI Risks

### Best-fit taxonomy: Yampolskiy (2016)

We chose Yampolskiy's (2016) *Taxonomy of pathways to dangerous AI* as our initial best-fit framework for developing a causal taxonomy for AI risk. We selected Yampolskiy's taxonomy as it was highly cited (116 citations, fifth most highly cited from the set of identified papers), simple, comprehensive, and provided sufficient definitions for each category.

Yampolskiy's taxonomy systematically classifies the ways in which an AI system might become dangerous based on two main factors:

1. Timing - whether the AI became dangerous at the pre-deployment or post-deployment stage, and
2. Cause - whether the danger arose from External Causes (On Purpose, By Mistake, Environment) or Internal Causes originating from the AI system itself (Independently).

Yampolskiy proposes that this taxonomy covers scenarios ranging from AI being purposely designed to be dangerous, to becoming dangerous by accident during development or after deployment, to turning dangerous due to environmental factors outside its control, or evolving to become dangerous through recursive self-improvement. Each "pathway" represents a set of causal conditions that lead to AI causing harm; for example, a person using a large language model (LLM) to generate fake news for political gain would be classified under Path B ("Timing: post-deployment; External cause: on purpose").

### Coding and iteration process

We started by using Yampolskiy's taxonomy to categorize a sample of risks from our database. We then identified themes in the AI Risk Database that didn't fit into Yampolskiy's taxonomy. We updated the taxonomy categories, criteria, and descriptions, then coded a further sample of risks. This process was repeated over three iterations until the taxonomy categories, criteria, and descriptions were stable. We describe this iteration process in detail in Supplemental Notes S3.

### Final taxonomy

The final version of the taxonomy, which we named the ***Causal Taxonomy of AI Risks***, included three categories of causal factors that specify how, why, or when an AI risk might emerge. The first category, ***Entity***, classified the entity (e.g., AI system, Human) that was presented as causing the risk to occur due to a decision or action taken by that entity. The second category, ***Intent***, classified whether the risk was presented as an expected outcome or unexpected outcome of an entity pursuing a goal. The third category, ***Timing***, classified the stage in the AI lifecycle that the risk is presented as occurring (e.g., pre-deployment, post-deployment). Each of these categories includes a third option, "Other," which captures risks that are not clearly categorizable within the primary options. Each of the categories is therefore mutually exclusive; each risk is classified under only one option within each category. The Causal Taxonomy is presented and described in more detail in the Results section.

## Development of Domain Taxonomy of AI Risks

### Best-fit taxonomy: Weidinger (2022)

We chose Weidinger et al (2022) "Taxonomy of Risks posed by Language Models" as our initial best-fit framework of consequences because it and its related papers [57,59] are among the highest cited in our review. Although this taxonomy was focused on language models, its set of categories was one of the most comprehensive, and it has been iterated upon over several publications. It included six areas of risks from language models:

1. Discrimination, Hate speech and Exclusion;
2. Information Hazards;
3. Misinformation Harms;
4. Malicious Uses;
5. Human-computer interaction Harms; and
6. Environmental and Socioeconomic Harms.

Each area of risk described several subcategories of risk.

### Coding and iteration process

We applied this taxonomy by coding as many of the included risks as possible using the Weidinger (2022) taxonomy. We operationalized the taxonomy by using the definitions or descriptions for each category from Weidinger (2022). Because several similar taxonomies were included in the set identified by the systematic literature review [26,57,59], we considered descriptions and definitions from any of these taxonomies in our initial coding.

We iterated on the taxonomy to accommodate risks that could not be coded against the existing Weidinger (2022) taxonomy. The most common risks that could not be accommodated were those related to AI system safety, failures and limitations; AI system security vulnerabilities and attacks; and competitive dynamics or other failures of governance to manage the development and deployment of AI systems. We describe this iteration process in detail in Supplemental Notes S4.

### Final taxonomy: Domain Taxonomy of AI Risks

The final version of the taxonomy, which we named the Domain Taxonomy of AI Risks, included seven domains of AI risk, and 24 subdomains of hazards and harms associated with AI. The domains were:

1. *Discrimination & toxicity*,
2. *Privacy & security*,
3. *Misinformation*,
4. *Malicious actors & misuse*,
5. *Human-computer interaction*,
6. *Socioeconomic & environmental harms*, and
7. *AI system safety, failures & limitations*.

As with Weidinger’s (2022) taxonomy, these risk domains are not mutually exclusive; many risks span multiple domains or subdomains due to their interconnected nature. For example, a risk related to AI-generated disinformation could be relevant to both the *Misinformation* domain and the *Malicious actors & misuse* domain. The Domain Taxonomy is presented and detailed in the Results section.

# Coding

Three authors were involved in coding risks against our taxonomies. Risks were coded by a single reviewer and discussed with the team where relevant. The coding process involved systematically categorizing each extracted risk according to the definitions within the relevant taxonomy. Based on grounded theory recommendations (cf. Charmaz, 2006; Corbin & Strauss, 2014), we coded risks as they were presented by the authors, aiming to capture the studied phenomena directly rather than impose our own interpretations or infer intent. When coding risks for our Causal Taxonomy, we categorized risks relevant to multiple levels of each causal factor (e.g., both pre-deployment and post-deployment) as "Other." In our Domain Taxonomy, we categorized risks relevant to multiple domains and subdomains (e.g., AI-generated disinformation) in the single most relevant category.

# Ongoing expert consultation and updates

We maintain the AI Risk Repository as a living resource through several mechanisms. First, we conduct biannual review cycles in which we screen documents proposed through our public submission form, author outreach, and expert consultations. That is, we solicit new or missed frameworks from the authors of existing frameworks, from interactions with other experts, and from a public form on our website. After screening proposed documents against our criteria, we extract and code risks using the procedures described above. We version the repository with each update logged and dated on the website. For example, between May 2024 and March 2025, we received 44 document recommendations and included 22 that met eligibility criteria, demonstrating active maintenance. Between March 2025 and December 2025, we received 78 document recommendations and included 9 that met criteria.

We acknowledge that long-term sustainability depends on continued institutional support. The repository is currently maintained by researchers at MIT FutureTech and the University of Queensland. We have designed the repository infrastructure to minimise maintenance burden: the public submission form, standardised extraction templates, and documented coding procedures allow updates to proceed efficiently. Should our capacity for active maintenance change, we commit to clearly indicating on the website whether the repository remains actively updated or has become a static archive.

# Resource Availability

## Lead Contact

Further information and requests for resources should be directed to and will be fulfilled by corresponding author, Michael Noetel (m.noetel@uq.edu.au).

## Materials Availability

This study did not generate new unique reagents or materials.

## Data and Code Availability

All data is publicly available at https://doi.org/10.17605/OSF.IO/CET8G.[25]

# Acknowledgments

This work was funded by Coefficient Giving (formerly known as Open Philanthropy), who had no role in the design, collection, analysis, interpretation or reporting of the data.

# Declaration of Interests

James Dao and Soroush Pour are employees of Harmony Intelligence, a company that conducts evaluations of AI risks. Their potential conflict of interest did not influence study design, risk classifications, or conclusions.

# Declaration of Generative AI and AI-assisted technologies in the writing process

Generative AI (Anthropic's Claude) was used to provide feedback on the writing in this manuscript. After using this tool, the authors reviewed and edited the content as needed and take full responsibility for the content of the published article.

## Author Contributions

Conceptualization: PS, AS, EG, MN, NT; Methodology: PS, AS, EG, JG, MN; Investigation and Data Collection: PS, AS, EG, JG; Analysis: PS, AS, EG, JG; Writing - original draft preparation: PS, AS, MN; Writing - review and editing: All authors; Project administration: AS, MN, NT; Supervision: MN, NT, RU, JD, SP, SC

# Supplementary Table S1. 20 most cited documents that present a taxonomy or classification of AI risks

Supplementary Table S1 presents the included taxonomies by citation rate. The included documents contained a range of highly cited taxonomies.

| Title | First author | First author affiliation (country) | Year | Type | Citations^ | Citations^ / year |
|---|---|---|---|---|---|---|
| Ethical and social risks of harm from language models | Weidinger | Deepmind (UK) | 2021 | Preprint | 1106 | 277 |
| Generative AI and ChatGPT: Applications, Challenges, and AI-Human Collaboration | Nah | City University of Hong Kong (China) | 2023 | Journal Article | 933 | 467 |
| Taxonomy of Risks posed by Language Models | Weidinger | Deepmind (UK) | 2022 | Conference Paper | 668 | 223 |
| The ethics of ChatGPT -- Exploring the ethical issues of an emerging technology | Stahl | University of Nottingham (UK) | 2024 | Journal Article | 352 | 352 |
| Trustworthy LLMs: a Survey and Guideline for Evaluating Large Language Models' Alignment | Liu | ByteDance Research (China) | 2024 | Preprint | 318 | 318 |
| The Dark Sides of Artificial Intelligence: An Integrated AI Governance Framework for Public Administration | Wirtz | German University of Administrative Sciences Speyer (Germany) | 2020 | Journal Article | 310 | 62 |
| AI Alignment: A Comprehensive Survey | Ji | Peking University (China) | 2023 | Preprint | 267 | 134 |
| Artificial Intelligence Trust, Risk and Security Management (AI TRiSM): Frameworks, Applications, Challenges and Future Research Directions | Habbal | Karabuk University (Turkiye) | 2024 | Journal Article | 226 | 226 |
| An Overview of Catastrophic AI Risks | Hendrycks | Center for AI Safety (USA) | 2023 | Preprint | 225 | 113 |
| The risks associated with Artificial General Intelligence: A systematic review | McLean | University of the Sunshine Coast (Australia) | 2023 | Journal Article | 199 | 100 |
| Sociotechnical Harms of Algorithmic Systems: Scoping a Taxonomy for Harm Reduction | Shelby | JusTech Lab, Google Research (USA) | 2023 | Conference Paper | 194 | 97 |
| Model Evaluation for Extreme Risks | Shevlane | Google Deepmind (UK) | 2023 | Preprint | 173 | 87 |
| GenAI against humanity: nefarious applications of generative artificial intelligence and large language models | Ferrara | University of Southern California (USA) | 2023 | Journal Article | 169 | 85 |
| SafetyBench: Evaluating the Safety of Large Language Models with Multiple Choice Questions | Zhang | Tsinghua University (China) | 2023 | Preprint | 164 | 82 |

| Title | First author | First author affiliation (country) | Year | Type | Citations^ | Citations^ / year |
|---|---|---|---|---|---|---|
| AGI Safety Literature Review | Everitt | Australian National University (Australia) | 2018 | Preprint | 164 | 23 |
| Sociotechnical Safety Evaluation of Generative AI Systems | Weidinger | Google Deepmind (UK) | 2023 | Preprint | 139 | 70 |
| Taxonomy of Pathways to Dangerous Artificial Intelligence | Yampolskiy | University of Louisville (USA) | 2016 | Journal Article | 138 | 15 |
| Safety Assessment of Chinese Large Language Models | Sun | Tsinghua University (China) | 2023 | Preprint | 130 | 65 |
| Evaluating the Social Impact of Generative AI Systems in Systems and Society | Solaiman | Hugging Face (USA) | 2023 | Preprint | 126 | 63 |
| Governance of artificial intelligence: A risk and guideline-based integrative framework | Wirtz | German University of Administrative Sciences Speyer (Germany) | 2022 | Journal Article | 122 | 41 |

*Note. ^ collected from Google Scholar on 21st November 2025. Seven organizational/industry reports [S1–3] were not indexed on Google Scholar and are therefore not listed.*

# Supplemental Note S1: Causal Taxonomy of AI Risk

The ***Entity*** variable captures which, if any, entity is *presented* as the main cause of the risk. It includes three levels: AI, Human, and Other. When the risk is attributed to AI, it means that the risk arises from decisions or actions made by the AI system itself, such as generating harmful content or disempowering humans. Conversely, when humans are seen as the source, the risks are implied to be due to human actions like choosing poor training data, intentional malicious design, or improper use of AI systems. The "Other" category captures cases where the focal entity is not a human or AI or is ambiguous. For example, “The software development toolchain of LLMs is complex and could bring threats to the developed LLM,” implies that the toolchain could be exploited by humans or AI.

The ***Intent*** variable captures whether the risk is *presented* as occurring as an expected or unexpected outcome from pursuing a goal. This variable has three levels: Intentional, Unintentional, and Other. Intentional risks are those that occur as expected outcomes from pursuing a specific goal, such as a case where AI is intentionally programmed to act deceptively or to exhibit bias. Unintentional risks reflect unintended consequences, such as an AI system inadvertently developing biases due to incomplete training data. The "Other" category captures risks where the intent is not clearly specified; for example, “The external tools (e.g., web APIs) present trustworthiness and privacy issues to LLM-based applications.” This includes cases where the risk may occur intentionally and unintentionally, such as "The potential for the AI system to infringe upon individuals' rights to privacy, through the data it collects, how it processes that data, or the conclusions it draws."

The ***Timing*** variable captures the stage in the AI lifecycle at which the risk is presented as occurring. The levels within this variable include Pre-deployment, Post-deployment, and Other. Pre-deployment risks are those that arise before the AI system is fully developed and put into use, such as vulnerabilities in the model due to coding errors. Post-deployment risks arise after the AI has been deployed, including issues like the misuse of AI for harmful purposes. Deployment is not defined in Yampolskiy (2016); we therefore interpreted it to mean when a product is being used by end users rather than just by developers. The "Other" category is used for risks that do not have a clearly defined time of occurrence (e.g., ”Resilience against adversarial attacks and distribution shift”). This includes cases where the presented risk may occur both before and after deployment; for example, "Generative models are known for their substantial energy requirements, necessitating significant amounts of electricity, cooling water, and hardware containing rare metals.”

## Supplementary Table S2. AI Risk Database coded with causal taxonomy: entity, intent, timing

Table S2 shows how the risks were coded against each category of causal factors. A majority of the risks were presented by authors of the documents as due to a decision or action by an artificial intelligence system (42%). Risks were presented equally as Unintentional (35%) compared to intentional (35%). Most of the risks were presented as occurring post-deployment (62%).

| Category | Level | Proportion |
|---|---|---|
| Entity | Human | 38% |
| | AI | 42% |
| | Other | 20% |
| Intent | Intentional | 35% |
| | Unintentional | 35% |
| | Other | 30% |
| Timing | Pre-deployment | 13% |
| | Post-deployment | 62% |
| | Other | 25% |

*Note. Totals may not match due to rounding.*

## Supplementary Table S3. AI Risk Database coded with Causal Taxonomy: entity x intent x timing

Table S3 shows how the risks intersect across our three causal factors. The most common triads of causal conditions under which an AI risk was presented as occurring were Entity = Human, Intention = Intentional, Timing = Post-deployment (18% of all risks). This was followed by Entity = AI, Intention = Unintentional, Timing = Post-deployment (14% of all risks).

| | | Intent | | |
|---|---|---|---|---|
| **Timing** | **Entity** | *Intentional* | *Unintentional* | *Other* |
| *Pre-deployment* | *Human* | 2% | 3% | • |
| | *AI* | • | 2% | • |
| | *Other* | • | • | • |
| *Post-deployment* | *Human* | 18% | 4% | 3% |
| | *AI* | 5% | 14% | 9% |
| | *Other* | 2% | 2% | 4% |
| *Other* | *Human* | 3% | 2% | 2% |
| | *AI* | 3% | 4% | 3% |
| | *Other* | • | 2% | 8% |

*Note. Taxonomy categories with a prevalence ≥ 10% are highlighted. Categories with a prevalence less than 2% in the AI Risk Database are shown as • for ease of interpretation.*

# Supplementary Table S4. Included documents coded with causal taxonomy

Papers varied significantly in terms of which causal factors they examined. Two documents have blank rows in the table because they did not present risks that could be coded against the Causal Taxonomy [S#26, 3,S#36, 4]. Human-related risks were identified in 65 out of 74 documents (85%), AI-related risks in 70 out of 74 documents (93%), and other risks in 58 out of 74 documents (77%). Regarding intent, intentional risks were noted in 65 out of 74 documents (85%), unintentional risks in 64 out of 74 documents (85%), and other intent-related risks in 67 out of 74 documents (89%). In terms of timing, pre-deployment risks were identified in 46 out of 74 documents (59%), post-deployment risks in 69 out of 74 documents (92%), and other timing-related risks in 63 out of 74 documents (84%). The majority of documents recognized risks arising from both human and AI actions, with near equal acknowledgment of intentional and unintentional risks. Post-deployment risks were more frequently discussed than pre-deployment risks, indicating the documents included focused more on the consequences of deployed AI systems.

| | | **Entity** | | | **Intent** | | | **Timing** | | |
|---|---|---|---|---|---|---|---|---|---|---|
| ID | First Author (Year) | *Human* | *AI* | *Other* | *Intentional* | *Unintentional* | *Other* | *Pre-deployment* | *Post-deployment* | *Other* |
| 1 | Critch (2023) [S5] | X | X | | X | X | | | X | X |
| 2 | Cui (2024) [S6] | X | X | X | X | X | X | X | X | X |
| 3 | Cunha (2023) [S7] | X | X | X | X | X | X | | X | X |
| 4 | Deng (2023) [S8] | | X | X | X | X | X | X | X | |
| 5 | Hagendorff (2024) [S9] | X | X | X | X | X | X | X | X | X |
| 6 | Hogenhout (2021) [S10] | X | X | | X | X | X | X | X | X |
| 7 | Kilian (2023) [S11] | X | X | X | X | X | X | | X | X |
| 8 | McLean (2023) [S12] | X | X | X | | | X | X | X | X |
| 9 | Meek (2016) [S13] | X | X | X | X | X | X | | X | X |
| 10 | Paes (2023) [S14] | X | X | X | X | X | | | X | |
| 11 | Shelby (2023) [S15] | X | X | X | X | X | X | | X | X |
| 12 | Sherman (2023) [S16] | X | X | X | X | X | X | | X | X |
| 13 | Solaiman (2023) [S17] | X | X | X | X | X | X | X | X | X |
| 14 | Steimers (2022) [S18] | | X | X | | X | X | | X | X |
| 15 | Tan (2022) [S19] | X | X | X | X | X | X | X | X | X |
| 16 | Weidinger (2022) [S20] | X | X | X | X | X | X | | X | X |
| 17 | Weidinger (2021) [S21] | X | X | X | X | X | X | | X | X |
| 18 | Weidinger (2023) [S22] | X | X | X | X | X | X | X | X | X |
| 19 | Wirtz (2022) [S23] | X | X | X | X | X | X | X | X | X |
| 20 | Wirtz (2020) [S24] | X | X | X | X | X | X | | X | X |
| 21 | Zhang (2022) [S25] | X | X | X | X | X | | X | | X |

| | | Entity | | | Intent | | | Timing | | |
|---|---|---|---|---|---|---|---|---|---|---|
| ID | First Author (Year) | *Human* | *AI* | *Other* | *Intentional* | *Unintentional* | *Other* | *Pre-deployment* | *Post-deployment* | *Other* |
| 22 | Hendrycks (2023) [S26] | X | X | | X | X | X | X | X | X |
| 23 | Vidgen (2024) [S27] | | X | | | | X | | X | |
| 24 | Gabriel (2024) [S28] | X | X | X | X | X | X | X | X | X |
| 25 | Shevlane (2023) [S29] | | X | | X | | | X | X | X |
| 26 | AIVerify (2023) [S3] | | | | | | | | | |
| 27 | Sun (2023) [S30] | X | X | | X | | X | | X | |
| 28 | Zhang (2023) [S31] | | X | | | X | X | | X | |
| 29 | Habbal (2024) [S32] | X | X | | X | X | | | X | |
| 30 | Liu (2024) [S33] | X | X | | X | X | X | X | X | X |
| 31 | EPIC (2023) [S1] | X | X | X | X | X | X | X | X | X |
| 32 | Stahl (2024) [S34] | | X | | | | X | | | X |
| 33 | Nah (2023) [S35] | X | X | X | X | X | X | X | X | X |
| 34 | Ji (2023) [S36] | X | X | X | X | X | X | X | X | X |
| 35 | Hendrycks (2022) [S37] | X | X | X | X | X | X | X | X | X |
| 36 | Sharma (2024) [S4] | | | | | | | | | |
| 37 | Giarmoleo (2024) [S38] | X | | X | X | | X | X | | X |
| 38 | Kumar (2023) [S39] | | X | | | X | X | | X | X |
| 39 | Saghiri (2022) [S40] | X | X | | X | X | X | X | X | X |
| 40 | Yampolskiy (2016) [S41] | X | X | X | X | X | X | X | X | |
| 41 | Allianz (2018) [S2] | X | X | X | X | X | X | | X | |
| 42 | Teixeira (2022) [S42] | X | X | X | X | X | X | X | X | X |
| 43 | InfoComm (2023) [S43] | X | X | | X | X | X | | | X |
| 44 | Coghlan (2023) [S44] | X | X | X | X | X | X | X | X | X |
| 45 | TC260 (2024) [S45] | X | X | X | X | X | X | X | X | X |
| 46 | Ferrara (2023) [S46] | X | X | X | X | | X | | X | X |
| 47 | G'sell (2024) [S47] | X | X | X | X | X | X | X | X | X |
| 48 | NIST (2024) [S48] | X | X | X | X | X | X | X | X | X |
| 49 | Bengio (2024) [S49] | X | X | X | X | X | X | | X | X |
| 50 | Zeng (2024) [S50] | X | X | X | X | X | X | | X | X |
| 51 | Everitt (2018) [S51] | X | X | X | X | X | X | X | X | X |
| 52 | Maham (2023) [S52] | X | X | X | X | X | X | X | X | X |
| 53 | Maas (2023) [S53] | X | X | X | X | X | X | X | X | X |
| 54 | Leech (2024) [S54] | X | X | X | X | X | X | X | X | X |
| 55 | Clarke (2022) [S55] | X | X | X | X | X | X | X | X | X |
| 56 | GOS (2023) [S56] | X | X | X | X | X | X | X | X | X |
| 57 | Ghosh (2025) [S57] | | X | | | | X | | X | |
| 58 | Abercrombie (2024) [S58] | X | X | X | X | X | X | | X | X |

| | | Entity | | | Intent | | | Timing | | |
|---|---|---|---|---|---|---|---|---|---|---|
| ID | First Author (Year) | *Human* | *AI* | *Other* | *Intentional* | *Unintentional* | *Other* | *Pre-deployment* | *Post-deployment* | *Other* |
| 59 | Schnitzer (2024) [S59] | X | X | X | X | X | X | X | X | X |
| 60 | Bengio (2025) [S60] | X | X | X | X | X | X | X | X | X |
| 61 | Uuk (2025) [S61] | X | X | X | X | X | X | X | X | X |
| 62 | Gipiškis (2024) [S62] | X | X | X | X | X | X | X | X | X |
| 63 | Hammond (2025) [S63] | X | X | X | X | X | X | X | X | X |
| 64 | Marchal (2024) [S64] | X | | | X | | | X | X | X |
| 65 | IBM (2025) [S65] | X | X | X | | X | X | X | X | |
| 66 | Li (2025) [S66] | X | X | X | X | X | X | X | X | X |
| 67 | DSIT (2023) [S67] | X | X | X | X | X | X | | X | X |
| 68 | Chin (2025) [S68] | X | X | X | X | X | X | X | X | X |
| 69 | Stanley (2024) [S69] | | X | X | | X | X | | X | X |
| 70 | Perlo (2025) [S70] | X | X | X | X | X | X | | X | X |
| 71 | Tang (2025) [S71] | X | | X | X | X | X | | X | X |
| 72 | Tse (2025) [S72] | X | X | X | X | X | X | X | X | X |
| 73 | Anwar (2024) [S73] | X | X | X | X | X | X | X | X | X |
| 74 | Wang (2025) [S74] | X | X | X | X | X | X | X | X | X |

# Supplemental Note S2: Detailed descriptions of domains of AI risks

## Domain 1: Discrimination and toxicity

***1.1 Unfair discrimination and misrepresentation.*** Humans hold inaccurate and overgeneralized beliefs about the characteristics, behaviors, and attributes of members of certain social groups. These stereotypical beliefs and the behavior that follows from them can misrepresent, exclude, demean, and disadvantage the individuals to whom they apply, reinforcing existing inequality. Human belief and behaviors shape every part of the design, development, and deployment of AI. Humans program AI systems, provide training data, and decide how data is processed and stored [S23]. As a result, AI models can encode associations that promote and amplify biased or discriminatory beliefs and behaviors. In decision systems, erroneous associations can systematically disadvantage certain groups. This may result in harmful decisions such as wrongful rejection of loan or mortgage applications [S15,S24], discriminatory hiring practices that exclude qualified candidates [S15,S24,S39], or the misidentification and unjust arrest of individuals in law enforcement contexts [S14,S39]. In text and image models, biased inputs can manifest in outputs that reinforce harmful stereotypes and prejudices that paint certain groups and individuals "... as lower status and less deserving of respect" [S15].

***1.2 Exposure to toxic content.*** Certain types of content have the potential to cause harm to the people who are exposed to them. These harms can vary in impact from minor (e.g., a transient experience of discomfort) to more severe (e.g., psychological, social, or physical consequences that are significant and/or enduring). Harmful speech is prevalent on the internet, particularly on social media platforms [S75]. Because AI models are commonly trained on vast amounts of internet data, they can internalize and regenerate these speech patterns in their output. In the context of LLMs, this output is known as "toxic content," an umbrella term that includes harmful, abusive, unsafe, and offensive material that violates community standards [S15,S43]. Frequently observed categories include content that promotes or encourages unlawful activities, hate, extremism, and violence [S6,S9,S22,S27]; provides hazardous or misleading high-risk advice [S9,S27]; or contains unwelcome or profoundly offensive, explicit material such as profanity, pornography, or child sexual abuse imagery [S22,S33].

***1.3 Unequal performance across groups.*** Decisions made during the development of an algorithmic system and the content, quality, and diversity of the training data can significantly impact which people and experiences the system can effectively understand, represent, and accommodate [S15,S17]. Biases and limitations introduced through these factors can lead to models that perform significantly worse for certain subpopulations compared with others, especially those defined by disability, gender identity, race, social status, and ethnicity [S15,S33]. For example, when LLMs are trained on a small number of languages, they can underperform for others [S20,S21]. The underperformance of algorithmic systems for certain groups may lead to a range of negative consequences such as the reduced ability or complete inability to use and benefit from the system [S15]; increased effort or challenges in using it effectively [S15]; feelings of alienation, frustration, and exclusion due to the lack of inclusive design [S15]; and ultimately, unequal outcomes across various domains [S15,S17].

## Domain 2: Privacy & security

***2.1 Compromise of privacy by obtaining, leaking, or correctly inferring sensitive information.*** In the context of generative AI, privacy violations arise when systems collect and divulge sensitive information that individuals or corporations do not consent to sharing with others [S6,S9,S16,S18,S22,S27]. Privacy violations can occur both accidentally and intentionally.

Examples of accidental causes include AI models that memorize and inadvertently reproduce or leak sensitive personal information present in their training data, such as names, addresses, and medical records [S6,S8,S9,S15,S21]. Even when personal data is not included in the training dataset or directly offered by the user, models can make inferences about sensitive or protected traits of individuals based on predictive correlations within their history of interactions [S7,S21], build profiles of users [S10,S15], or train AI systems [S32]. As a result, models may save and reproduce sensitive information derived from prior interactions, such as classified intellectual property [S7,S21]. A notable example is the case where Samsung employees accidentally leaked confidential intellectual property to OpenAI after using ChatGPT to help with coding tasks [S7,S21].

Intentional causes include the malicious design and use of AI to exploit users' trust by influencing them to share personal or private information about themselves or others [S28]. Privacy attacks, such as membership inference, could allow adversaries to gain knowledge of the private records used to train an AI model [S28]. Malicious actors could also deliberately extract private information from a model by crafting prompts designed to exploit the model's knowledge of sensitive data [S6].

***2.2 AI system security vulnerabilities and attacks.*** AI systems, like other software systems, face a range of security threats. These issues may arise from inherent weaknesses in the design of AI algorithms, the data used to train the models, or the operational context. Specific examples include:

- *Toolchain and dependency vulnerabilities* that arise unintentionally through the use of automated code-generation tools (e.g., Github Copilot, Python language, OpenCV), deep learning frameworks (e.g., Tensorflow, PyTorch), or as a result of complex interdependencies in the development environment [S6].
- *External tool and API integration into AI system applications* can compromise the trustworthiness and privacy of systems due to their potential unreliability or susceptibility to adversarial control [S6].
- *Security vulnerabilities in physical and network infrastructure,* such as vulnerabilities in graphics processing units, or GPUs, or to sophisticated attacks like side-channel and rowhammer attacks, can lead to unauthorized access or manipulation of model parameters when used during training of AI systems [S6]. The use of distributed network systems for training AI systems such as LLMs exposes them to network-specific threats like pulsating attacks or congestion.
- *Direct manipulation of AI systems* such as adversarial attacks and instruction-based attacks. Adversarial attacks focus on altering the model's learning process or extracting its data. They include perturbations designed to deceive models into incorrect outputs, extraction attacks to steal model insights, and poisoning attacks to alter model behavior [S6,S28,S33]. Instruction-based attacks manipulate the way the model handles and responds to inputs [S9,S30,S33]. Attackers deliberately craft prompts to induce models to produce biased or unsafe outputs (a.k.a. ‘jailbreaking’). This manipulation directly targets the operational aspects of AI systems with the intent to cause harm.

## Domain 3: Misinformation

***3.1 False or misleading information.*** LLMs can sometimes generate content that is factually incorrect, misleading, poorly researched, or unintelligible [S1,S7–9,S15,S35]. Risks in this category occur accidentally and not as a result of humans intentionally trying to cause harm, as is the case with *disinformation [S20,S33]*. Common sources of AI misinformation include noisy training data [S6,S33], sampling strategies that introduce randomness [S6], outdated knowledge bases [S33], and fine-tuning processes that encourage sycophantic behavior [S6]. Incorrect and misleading information generated by LLMs can result in a range of actual and anticipated negative outcomes. Individuals exposed to false information may form inaccurate beliefs and perceptions. This undermines their autonomy and ability to make free and informed choices [S20,S21]. Where inaccuracies in LLM predictions influence an individual’s decisions and actions, the individual may experience indirect physical, emotional, or material harms [S20,S28] especially but not exclusively in high-stakes domains such as mental health [S30,S31], physical health [S30,S31,S35], law [S20], and finance [S27]. For example, an LLM that offers misleading information about medical drug use may cause a consumer to harm themselves or others [S30].

***3.2 Pollution of information ecosystems and loss of consensus reality.*** This subcategory covers the diverse effects of AI-driven personalisation and content-generation technologies on the information landscape. As AI systems become more adept at tailoring content to individual preferences, they risk creating “filter bubbles” [S10]. These are informational cocoons where individuals are predominantly exposed to news and opinions that align with their pre-existing beliefs. AI-driven filter bubbles are likely to be more pervasive and intense than those driven by traditional internet browsing and recommendation algorithms: They adapt to individual preferences in a more sophisticated manner (e.g., through reinforcement learning and analysis of user behavioral data) [S28], integrate seamlessly into daily life, and are more opaque. An overreliance on hyper-personalized AI information sources could lead to a “splintering” of shared reality, where different groups of people have vastly different understandings of what is true or important [S10,S28]. This is likely to be exacerbated by the proliferation of AI-enabled content generation technologies that spread misinformation at higher rates (e.g., clickbait), potentially making consumers generally distrustful of information and important institutions [S1,S28,S37]. A shared sense of reality is fundamental to social solidarity. Where societal bonds are weakened, individuals may become more hostile towards opposing views. This can hinder constructive dialogue on critical collective issues like climate change and public health [S28].

## Domain 4: Malicious actors and misuse

***4.1 Disinformation, surveillance, and influence at scale.*** Advances in AI have made powerful dual-use technologies like voice cloning, deep fakes, content generation, and data-gathering tools cheaper, more efficient, and easier to use [S7]. With modest hardware requirements, these technologies are now within the reach of a broader group of users, including those with malicious intent. Disinformation is already a serious issue [S26] and involves the deliberate propagation of false or misleading information, usually with the intent to cause harm, influence behavior, or achieve a financial or political advantage [S1,S43].

AI tools could be used to amplify the impact and scope of disinformation through more personalized, convincing, and far-reaching messaging [S20,S21,S24,S26,S28]. For example, the use of advanced AI in phishing schemes enables cybercriminals to automate the creation of highly sophisticated image, video, and audio communications [S7,S28,S32]. These communications can be tailored to individual recipients (sometimes including the cloned voice of a loved one), making them more likely to be successful and harder for both users and anti-phishing tools to detect [S28]. In the realm of surveillance, AI could support and enhance the mass gathering of personal data [S20,S21,S28]. Historically, mass surveillance required extensive manual effort. Machine learning tools can now link and process large datasets much more efficiently and cheaply than human analysts and can make predictions and decisions without human intervention [S20]. Through microtargeting, actors could manipulate individual behavior more subtly and effectively using AI-derived insights from their personal data and online behavior.

In the hands of nefarious state actors, such capabilities could be used to enhance the effectiveness of illegitimate domestic surveillance campaigns and to facilitate oppression and control [S28]. All of the capabilities mentioned above could converge to facilitate the large-scale manipulation and control of what people see, hear, and believe. A form of this is automated censorship in which AI systems are used to selectively suppress or block specific types of information, content, or voices deemed undesirable to those controlling the AI [S23]. AI can not only be used to silence voices but also to entrench specific agendas: Actors (be they political figures, organizations, or state actors) could use AI to distribute incorrect information about electoral systems and processes [S27], produce persuasive propaganda [S2,S33], and systematically exert control over public opinion and political debates on a large scale [S23]. During the 2016 Brexit referendum, a network of over 10,000 AI-powered political bots were employed to distribute fake and hyperpartisan news [S2,S76]. The selective visibility of information can lead to the formation of incorrect or incomplete beliefs about what is happening in the world. This ability to shape public discourse can maintain or increase the power of those in control while keeping the public in the dark about critical issues that may affect their lives and their society.

***4.2 Cyberattacks, weapon development or use, and mass harm.*** AI may be used to gain a political or strategic advantage or to cause harm at scale through cyber operations or the development and use of weapons. Advancements in AI have provided malicious actors with powerful tools that can lead to more frequent, more severe, and more precise cyber attacks [S10,S11,S20,S23,S28,S33,S37]. Hackers could use the coding abilities of AI assistants to develop malicious malware more effectively and at lower cost [S2,S9,S10,S16,S20]. With AI, even those with limited coding and technical experience [S1,S28] could teach a model to produce and optimize malware code that discovers and exploits system vulnerabilities, including both self-replicating [S2,S43] and automated software [S6,S29]. The development and application of weapons could also be sped up and intensified through AI. For example, AIs with specialized knowledge of bioengineering could make it easier for more actors to design new bioweapons [S26,S43]. For example, in 2022 a small pharmaceutical company used generative AI to develop 40,000 chemical nerve agents in less than six hours [S77]. AI could also enable autonomous devices, such as drones, to be used as weapons [S2]. In fact, AI has already assisted in the development and application of Lethal Autonomous Weapons Systems (LAWS) – weapons that can operate without human oversight and use computer algorithms to identify and attack targets [S10,S32]. Autonomous weapons may fail in ways that other AI systems do, such as through a lack of capability, robustness, or loss of control, meaning that they would cause harm that was not intended by their developers or operators. In these circumstances, the risks from LAWS would not be limited to malicious actors or misuse. However, in most cases, we conceptualize risks from LAWS as relating to purposeful decisions made by humans controlling the weapons. AIs deployed by states in conflict could also be integrated in conventional defense or mass-casualty weapons [S27,S42]. These integrations could range from AI-controlled aerial combat to the operation of AI as part of a country's nuclear arsenal as a “fail-safe” mechanism [S37].

Overall, AI’s ability to process vast amounts of data quickly may empower actors to act on a much larger scale than would otherwise be possible. AI can manage multiple attack vectors simultaneously, coordinating them to maximize disruption and harm. Malicious actors may intentionally cause mass harm through terrorism or the disruption of law enforcement [S5]. For example, AI could automate the process of finding and exploiting vulnerabilities in software used by millions of people [S28]. AI could also be used to identify vulnerabilities in national power grids and strategically target key components to cause outages or to determine optimal release points for biological agents to maximize impact and spread.

***4.3 Fraud, scams, and targeted manipulation.*** AI capabilities have the potential to be exploited for personal gain at the expense of others via deception and manipulation. This can take various forms including cheating, fraud, scams, and the use of deepfakes for blackmail or humiliation. It is currently very difficult to distinguish human text from text that is AI-generated [S9]. This increases opportunities for cheating in settings where rewards depend on the communication of original thought. In academia, students may use AI to quickly generate essays or other coursework and claim it as their own [S6,S9,S35]. If students’ regularly and inappropriately rely on AI for their schooling, this could undermine academic integrity and genuine intellectual development [S9]. In science, researchers could use AI unscrupulously to produce professional outputs [S6]. If widely adopted, this practice could dilute the overall quality of scientific discourse [S9].

Generative AI products may also be used to increase the reach and potency of various dishonest schemes. Advanced AI assistants can produce HTML, CSS, and other web development languages, allowing for the rapid creation of convincing fraudulent websites and applications at scale [S28]. In the context of social media, generative adversarial networks (GANs) have been used to create images of human faces that look authentic [S28]. These images can be uploaded as profile pictures to fake accounts to make them seem more trustworthy. AI models can also be trained on speech or writing data from a specific individual. This allows the model to impersonate someone very convincingly without consent. Scammers could use this capability to request sensitive information or financial aid by pretending to be a trusted contact [S20,S21]. AI has recently advanced in generating realistic deep fakes which have enabled new forms of targeted harassment and extortion [S10]. A particularly damaging type of abuse facilitated by deep fakes involves creating non-consensual sexual imagery with the intent to cause a subject social injury or manipulate them into performing desired actions [S1,S15,S22,S33]. Even if a deep fake is exposed as inauthentic, it can continue to impact a person’s life in significant ways [S1] through the loss of job opportunities, social isolation, and ongoing harassment or defamation.

## Domain 5: Human-computer interaction

***5.1 Overreliance and unsafe use.*** Users may come to trust or rely on AI systems beyond their actual capabilities or to anthropomorphize AI systems, which can lead to emotional or material dependence and inappropriate relationships with or expectations of AI systems. Users who develop trust in an AI may be harmed if this trust is miscalibrated, such as relying on an AI to provide advice, make decisions, or otherwise act in complex, risky situations for which the AI is only superficially equipped [S28]. For example, a user experiencing a mental health crisis may request psychotherapy from an AI with whom they have formed a connection. Were the AI to respond with insensitive or destructive advice, this could put the person in immediate danger [S28].

When people interact with AIs that use convincing natural language, they may start to perceive them as having human-like attributes and invest undue confidence in their capabilities [S20,S21]. Anthropomorphic perceptions of AIs may encourage users to develop emotional trust in the systems [S9], which can make users more likely to follow suggestions, accept advice, and disclose personal information [S20,S21]. This trust could be exploited by manipulative actors who wish to harvest user's sensitive data or influence their decisions and actions for purposes which are unlikely to be in the user's best interests [S20]. For example, AI systems could be used to power increasingly manipulative recommendation algorithms [S22].

Beyond inappropriate trust, humans may develop broader and more vital attachments to AI systems that undermine their ability to function adaptively in the long term. For example, where an attachment becomes an uncontrolled dependence, a person's ability to make free and independent decisions could be compromised [S28]. More broadly, as AIs increasingly take over human tasks [S23] and become better at simulating satisfying and authentic interactions, people may increasingly withdraw from human relationships to immerse themselves in AI-mediated environments [S9,S28]. Over time, widespread preference for interacting with AIs could weaken social ties between humans. This shift could induce psychological distress because genuinely reciprocal relationships are often important to human satisfaction and well-being [S2,S23,S28].

***5.2 Loss of human agency and autonomy.*** As AI systems become increasingly capable and intelligent, humans may be tempted to delegate many of their decisions and actions to AI [S14]. Although such delegation can be beneficial (e.g., by saving time or money), it may lead to undesirable outcomes where unconstrained or inappropriate. For example, if AIs take over tasks that typically require human creativity and analytical thinking, humans may engage less frequently in these cognitive processes. Over time, this may lead to a decrease in our ability to think critically and solve problems independently [S35]. As individuals become more reliant on AI for everyday decisions – from what to eat and how to spend to more significant choices like career and relationships – there is a risk that they will lose their sense of free will and autonomy [S23,S28]. If AIs begin to shape a person's life path in ways that do not align with their original aspirations and desires, this could limit their personal growth and prevent the pursuit of a fulfilling life [S15,S28,S39]. At a societal level, organizations may hand over control to AI systems to stay competitive or reduce costs [S37]. If a significant number of organizations adopt AI systems and automate decision-making processes, especially in a way that is opaque and difficult to challenge, it could lead to widespread job displacement and a growing sense of helplessness among the general population [S10].

## Domain 6: Socioeconomic & environmental harms

***6.1 Power centralization and unfair distribution of benefits.*** Developing cutting-edge AI technologies requires significant computational power, expertise, financial resources, and datasets [S1,S10,S17,S28]. As such, there is a risk that the most influential and valuable AI technologies, along with their political and competitive benefits, could be monopolized by a handful of powerful entities, such as major technology corporations or governments [S10,S37]. If AI is primarily controlled by a few entities, its instructions and data could reflect their narrow perspectives, experiences, and priorities [S28,S38]. Without inputs from diverse parties, AI systems may operate in ways that systematically favor the controlling entity and fail to serve the needs of the broader population. Current AI systems suffer from global inequities in performance and access that disproportionately impact historically disadvantaged groups. These inequities often relate to language, culture, knowledge, paywalls, and access to hardware or the internet [S20–22,S28]. As the integration of AI systems into a wider range of applications and services becomes simpler, these existing disparities could be entrenched and broadened [S15,S20,S28,S35].

In situations where AI is embedded in essential services (e.g., social security and welfare, tax filing, insurance, hospital infrastructure), many more people, including those who are currently disenfranchised, may be denied appropriate access to critical resources and benefits [S28]. The centralization of AI systems and their authoritative power could also enable governments or other empowered actors to pursue overly aggressive forms of censorship, oppression, and surveillance [S17,S26]. Over time, these measures may become normalized, weakening or eliminating the checks and balances that prevent the abuse of power. These conditions may foster the development of a totalitarian regime [S2]. Once AI systems are deeply integrated into social control mechanisms, it may be extremely difficult to dismantle such a regime.

***6.2 Increased inequality and decline in employment quality.*** AI systems are increasingly automating many human tasks, potentially leading to significant job losses [S2,S14,S20,S35]. If AI is able to provide large-scale labor that is less expensive and more effective than human labor, it could take over major industries (e.g., manufacturing, crowdwork platforms, software engineering), causing mass unemployment [S9,S13,S23]. This displacement of labor could worsen existing social and economic inequalities [S1,S23], as those most vulnerable to automation are likely to currently occupy positions of disadvantage [S9,S20]. New disparities may also arise between those who are able to adapt their skills to complement AI systems and those who are not [S35]. Aside from the availability of jobs, AI automation may negatively impact job quality and security [S35]. The roles that remain after widespread automation could be more monotonous and less engaging as AI takes on more complex tasks [S1].

Furthermore, the threat of replacement by AI could result in exploitative dependencies between human workers and their employers. In order to remain competitive with faster, more knowledgeable AI assistants, human workers may be pressured to accept lower wages, fewer benefits, and poorer working conditions. This dynamic can be observed today: Generative AI companies have a history of exploiting dispensable workers (e.g., refugees, prisoners, low-income individuals) for crowdwork that is fraught and unfair [S1,S17,S22]. The future development of AI systems may continue to power such unfair disparities between AI companies and their workers [S1].

***6.3 Economic and cultural devaluation of human effort.*** Generative AI is trained on vast bodies of internet data, including text and images. Frequently, this data contains original, copyright-protected works that have been obtained without authorisation [S1,S9,S35]. This may present a risk to authors if users extract these works verbatim from the system's data [S9,S27,S33]. Relatedly, models may produce content that does not, in a strict sense, unlawfully copy an author's work but benefits substantially from its unique style, method, or genre [S6,S20–22,S28]. If models are able to produce synthetic replacements for such work at a speed and scale that surpasses humans, this may jeopardize the ability of creators to earn an income and stymie human innovation and creativity [S9,S20,S22]. A particularly damaging case of this may be developers using AIs to request off-the-shelf computer code [S7]. Although some authors are attempting to sue AI companies for the appropriation of their work [S7], this and similar issues fall into a legal “gray area” within which current frameworks do not offer a secure path to recourse [S1,S9].

Several synergistic risks arise from the widespread dissemination and use of AI-generated cultural products. Because AIs optimize for repeated patterns in their training data, it is possible that their works will lack the diversity and unpredictability often celebrated in human works [S35]. Where synthetic works are adopted on a large enough scale, this could homogenize cultural experiences. Similarly, AIs do not understand the contextual significance of the cultural elements that they use. If AI enables the extensive commodification of certain products, it may expropriate their cultural value [S20]. For example, an AI might use Australian Aboriginal or Torres Strait Islander artwork in its designs without acknowledging or respecting their symbolic meanings.

***6.4 Competitive dynamics.*** AI technology has the potential to redefine power dynamics across economic, political, and social spheres. As a result, many countries and corporations are investing heavily in AI research and development with the goal of becoming leaders in the area. While market competition can lead to beneficial economic and consumer outcomes, it also presents various risks, particularly in the field of AI [S26]. In intensely competitive markets, AI developers and deployers may have an incentive to prioritize short-term, internal goals (e.g., profit or influence) to “secure their positions and survive” [S26], at the expense of external goals that encourage longer-term societal well-being [S26]. A key concern is that AI companies may cut safety corners, releasing insecure and error-prone systems in a bid to stay ahead [S12]. These immature systems may present risks that are hard to identify and evaluate [S18]. Akin to the fossil fuel industry, profit-focused developers may allow their technologies to cause widespread externalities, such as “pollution, resource depletion, mental illness, misinformation, or injustice” [S5]. Countries or other state-like actors may engage in an AI-enabled military arms race, which could encourage the making of bad bets with a high potential for harm [S23,S26]. For example, they may give AI the autonomy to conduct cyberattacks, drone swarms, or disseminate propaganda and disinformation.

***6.5 Governance failure.*** Governance failure refers to the risks and harms that arise when institutional, regulatory, and policy mechanisms fall short of effectively managing and overseeing the development and deployment of AI systems. Several issues make robust AI governance challenging to implement [S35].

First, it is difficult to determine who is responsible or liable when AI systems fail or make decisions that result in negative consequences [S1,S2,S23,S40]. At present, there exists no comprehensive framework specifically designed to assign legal responsibility to AI agents [S13]. Traditional legal principles are based on human actors, whose intentions and actions can generally be identified and judged. AI’s decision-making, on the other hand, is often unpredictable, opaque, and involves complex interactions between millions of parameters [S35]. This complexity makes understanding how an AI arrived at a decision, and consequently who is responsible for the consequences of that decision, very difficult [S24]. In the absence of a regulatory or legal incentive to take safety engineering seriously, developers may release poorly designed AI systems [S13], and people harmed by those systems may be left without recourse [S42].

A second challenge for effective AI governance is the rapid pace at which AI systems evolve. Typical governance and policy processes are inherently slow. Developing, proposing, debating, and implementing new regulations often involves multiple stakeholders, including government bodies, industry experts, and consultations with the public. The mismatch between the speed of AI advancements and their regulation may result in immature regulations that overlook important aspects of AI governance [S23]. The "great scope and ubiquity" of AI increases the difficulty of comprehensive governance [S23]. At present, many emerging aspects of AI-generated content are not explicitly addressed in copyright laws [S35]. Regulatory lags such as this could become increasingly dangerous as AI systems develop more harmful capabilities.

A third challenge for effective governance is an inability to influence AI developers and deployers to take safe actions. Frequently, this inability is driven by an asymmetry of information between technology companies and regulators [S35]. Technology companies often have far better knowledge about the capabilities, functioning, and potential uses of their AI systems; they possess both the technical expertise and the proprietary data that inform AI development. Without access to this knowledge, regulators can find it difficult to craft targeted rules that address the specific challenges posed by AI.

***6.6 Environmental harm.*** Generative models, especially those that use deep learning techniques, require vast amounts of resources to train, test, and deploy [S9,S17]. Training a model can take days or weeks. This process requires powerful processors that consume large amounts of electricity and produce significant greenhouse emissions [S1,S9,S17,S20,S21,S40]. The hardware that runs AI models – primarily GPUs – often contains rare metals (e.g., nickel, cobalt, and lithium) that are costly and environmentally taxing to collect and process [S9,S14,S15]. Data centers that house models generate significant heat and require substantial water and energy to cool [S20]. Secondary environmental impacts include emissions from AI-enabled applications [S20]. The resource requirements of AIs can impose significant costs on the natural environment [S22,S34], as they are often acquired and used in ways that are unsustainable [S9] (i.e., produce significant carbon emissions), deplete resources, and damage built environments [S15].

## Domain 7: AI system safety, failures & limitations

***7.1 AI pursuing its own goals in conflict with human goals or values.*** Continued massive investment in AI research and development raises the possibility that AI systems could eventually rival or surpass human intelligence. AIs could cause permanent and severe harm when the objectives of human or superhuman-level AI are misaligned with human values and goals, and if they evade our control [S9,S41]. The literature has identified several technical challenges that may impede robust alignment, such as reward hacking, reward tampering, proxy-gaming, goal misgeneralisation, or goal drift [S9,S10,S26,S28,S36,S37]. The literature has also identified a range of harmful behaviors that AIs may exhibit if these misalignment challenges cannot be solved and if systems reach a certain level of advancement. For instance, misaligned AIs may resist human attempts to control or shut them down [S9,S22,S23,S28,S34,S40,S43]. In many cases, gaining more control or power (e.g., money, energy, resources) is an effective way for an AI to optimize its objectives [S9,S23,S28,S36,S37]. Absent strong behavioral constraints, a sufficiently advanced AI may act upon these drives.

Misaligned AIs may acquire, develop, or use dangerous capabilities to evade human control and oversight and to cause mass harm. Description of some of these capabilities are provided in subdomain *7.2 AI possessing dangerous capabilities that could cause mass harm*, and include situational awareness, cyber-offense, deception, persuasion and manipulation, weapons acquisition, strategic planning, and self-proliferation [S29]. For example, an AI system that possesses the dangerous capability of *situational awareness* may hold knowledge about its status as a model, how it is expected to operate in its surroundings, its ability to control these surroundings, and how people may respond to its behaviors [S28,S29]. A misaligned AI system could use information about whether it is being monitored or evaluated to maintain the appearance of alignment, while hiding misaligned objectives that it plans to pursue once deployed or sufficiently empowered [S9,S23,S28,S36,S37]. A misaligned AI system that possesses the dangerous capabilities of persuasion, manipulation, and/or deception may use these capabilities to coerce humans into taking harmful actions that they would not otherwise take [S22,S34], such as giving the AI system access to resources or weapons [S9,S23,S28]. Combinations of dangerous capabilities may be used by a misaligned AI system: *situational awareness* allows a system to detect when it can pursue its goals without being monitored, *deception* allows a system to mislead users about its behavior and goals; *persuasion or coercion* allows a system to influence users to provide it with resources; the resources can then be used for *self-improvement* and *self-replication* to resist attempts of shut down or control so that the system can pursue its goals [S9,S23,S28,S29,S43].

***7.2 AI possessing dangerous capabilities.*** AI systems may develop or acquire capabilities that can cause large-scale harm if used by humans, misaligned AI systems, or due to a failure in the AI system. These capabilities are described as dangerous because they can be used to threaten security or exercise control over humans. These capabilities may be intentionally designed into an AI system, may emerge unpredictably during development or training of a system, may be acquired by an AI system in its environment (e.g., through the use of tools), or be provided by a user [S29].

One example of a dangerous capability is *manipulation and persuasion*, where an AI system can convince humans to believe things that are irrational or false or to engage in dangerous behaviors [S28,S29]. An AI system, for instance, could convince people to transfer ownership of property or legal statuses to entities controlled by the AI or its user [S13]. Other dangerous capabilities include *political strategy* and *knowledge of social dynamics* that can be used to obtain and wield power [S29]. *Cyber-offense* skills may enable an AI system to gain ongoing unauthorized access to hardware, software, or data systems and work strategically towards a planned goal while minimizing the risk of detection [S29,S36]. AI systems could hack into control systems and military hardware, allowing it to commandeer weapons [S29]. Additionally, models may become capable of assisting in the *research and development of novel weapons*. In this circumstance, they may give a human collaborator step-by-step guidance on the creation of weapons [S29].

AI systems may also develop highly effective "evasion skills," such as *situational awareness* [S8,S12,S13,S25,S28–30,S36,S42,S43] and *deception [S29,S40,S43]*, which would allow them to outmaneuver human oversight and control. Situational awareness refers to the AI's ability to understand and interpret its environment and situation when it is being monitored, trained, or deployed, along with the location of its technical infrastructure. Deception refers to the model's ability to intentionally generate false or misleading statements that seem credible to humans, anticipate how these statements might influence feelings and decisions, and strategically conceal or offer information to sustain its credibility [S29]. AIs may also acquire a suite of capabilities necessary for *self-proliferation*. This could include skills to escape operational confines and evade detection, autonomously produce income, obtain server space or computational resources, and copy their underlying software and parameters [S29,S43]. Aside from self-proliferation, AIs may develop the ability to construct new dangerous models or alter current models to enhance their destructive capacity [S29,S43]. Finally, sophisticated AI systems may become capable of *strategic planning*, such as creating and executing intricate, long-term strategies that can adjust to changing conditions and that are effective across many different contexts, including novel or adversarial situations [S8,S12,S13,S28–30,S36,S42,S43]. The highest risk scenarios in this subcategory are likely to arise not from a single capability, but from the convergence of several capabilities [S29]**.**

Each of these dangerous capabilities may be used by an AI system to cause harm when intentionally directed by “legitimate” human actors (e.g., state intelligence or military agencies), or malicious human actors (e.g., criminals, terrorists), as described in the domain *4 Malicious actors & misuse*. However, dangerous capabilities may also help an AI system to pursue its goals, as described in *7.1 AI pursuing its own goals in conflict with human goals or values*. Instead of using these capabilities at the direction of a human, an AI system may employ dangerous capabilities to deceive or manipulate humans, gain resources, and evade shutdown or control. One scenario is that an AI system’s possession of dangerous capabilities may itself be a sufficient condition for the loss of control of an AI system [S12,S26].

***7.3 Lack of capability or robustness.*** This subcategory includes the broad set of risks associated with the failure of an AI system to fulfill its intended purpose. The literature identifies four main situations in which an AI may fail to perform as expected or desired.

First, the AI system can fail if it *lacks the inherent capability or skill required to perform a task* or if this skill is poorly developed [S10,S28,S41]. The consequences may be particularly harmful in situations where an AI is required to reason at a human level about important moral issues but does not possess this capability or possesses an obsolete or divergent version of it that is not aligned with human values [S8,S12,S13,S28,S30,S31,S36,S42,S43]. For example, an AI-based healthcare system tasked with prioritizing patient treatment schedules might be unable to appropriately consider ethical principles like justice and beneficence, leading to prioritizations that are technically effective but immoral. Cultural, individual, and temporal differences in ideas of what is “right” or “ethical” compound the challenge of endowing AI with appropriate and adaptable ethical standards that are fit for all purposes [S24].

Second, the AI system can fail when it is *not robust in “out of distribution (OOD)” situations:* data or conditions that were not anticipated during its training phase [S19,S25,S28,S33,S42,S43]. These failures may occur because the training data did not confer a particular skill to the AI [S35] or because the skill was learned in a fragile way that did not permit generalization to unpredictable and complex real-world environments [S18,S28].

Third, the AI system can fail or become unstable when it is *unfit to handle unusual changes or perturbations in input data* [S16,S19,S33]. These unusual changes could be due to environmental noise, invalid inputs, or adversarial inputs from a malicious attacker [S40,S43].

Fourth, the AI system can fail as a result of *oversights, undetected bugs, or errors in the design process* [S19,S41]. A common design oversight is a lack of comprehensive technical safeguards to prevent unintended downstream uses or consequences [S5,S19]. These factors can result in significant harms such as lab leaks or addictive products [S5]. Critical design choices about the algorithm, optimization techniques, and model architecture can also directly influence whether a system is able to consistently perform its intended function, leading to possible harm [S19].

***7.4 Lack of transparency or interpretability.*** Many AI models, especially those based on deep learning, involve complex mathematical structures that can be difficult to interpret, even for experts [S39]. AI systems are also often trained on vast datasets that they use to learn patterns and make predictions. The complexity and volume of this data mean that the learning process – how data points influence the AI's development and final decisions – can be opaque [S35,S40,S42]. Furthermore, in many cases, the algorithms, data, and specific methodologies used in developing AI are considered proprietary, and companies may be reluctant to share them openly [S16]. Because of these factors, obtaining understandable information about the decision-making process for AI can be challenging [S13]. This lack of transparency and interpretability raises issues for several stakeholders.

For users, an inability to interrogate how an output was obtained may lead to a lack of trust and confidence in the system's results and to resistance to adopting the technology [S10,S14,S33,S35,S39,S40]. Users may also misinterpret or struggle to find and amend errors in the model's results [S16,S35].

For regulators, AI opacity can frustrate auditing or other compliance standards [S10,S35]. For example, auditors faced with obscured or incomplete information about an AI system may find it difficult to check the system for biases, accuracy, and fairness or to reproduce it [S40,S42]. Where an AI system's compliance cannot be assessed, a "responsibility gap" may be created [S38], and it may become difficult or impossible to hold systems or relevant actors accountable for their actions [S16,S18,S39,S40,S42]. In certain sectors, decisions made by AI systems can have profound consequences. In healthcare, AI might be used to diagnose diseases or recommend treatments where incorrect decisions could directly affect patient outcomes. In the military, AI might be used in operations that could impact national security or lead to significant loss of life. In these areas, transparency and accountability of the AI system are particularly pressing issues [S40].

***7.5 AI welfare and rights.*** At a sufficient level of complexity, it is possible that AI systems could acquire the ability to have subjective experiences, particularly pleasure and pain. Some consciousness researchers and philosophers consider the possibility of sentient AI theoretically feasible [S78,S79]. Where AIs become sentient, they may deserve moral consideration and therefore a range of the rights currently afforded to many forms of human, animal, and environmental life [S13]. Systems may be mistreated or harmed if these rights are not implemented responsibly or we accidentally or intentionally treat AIs as non-sentient where they are sentient. As AI technology advances, it will become more challenging to assess whether an AI has developed the sentience, consciousness, or self-awareness that would grant it moral status.

***7.6 Multi-agent risks.*** AI systems that interact autonomously with each other will form multi-agent systems [S63]. Multi-agent systems are associated with unique risks beyond those posed by individual AI systems. These risks fall into three main failure modes depending on the objectives of the AI agent and how humans expect systems to behave:

- *Miscoordination* occurs when AI agents fail to cooperate effectively despite sharing the same goals. This can be caused by agents choosing *incompatible strategies* to achieve mutual ends. For example, driving models trained on United States vs Indian cultural conventions for yielding to emergency vehicles block traffic in 77.5% of scenarios despite their shared goal of clearing a path [S63].

- *Conflict* occurs when AI agents with different but overlapping goals compete in harmful ways. For example, by intensifying competition over shared resources or escalating military tensions. They could also make novel forms of conflict possible through more advanced and accessible methods of coercion and extortion.
- *Collusion* occurs when undesired cooperation emerges between AI agents, allowing them to circumvent safeguards or manipulate markets. For example, AI systems may be able to develop hidden communication channels without explicit training. [S80] show that advanced LLMs can covertly exchange *steganographic* messages undetected by equally capable oversight systems, using natural language cues and shared context. In market settings, AI systems may learn to collude because it is the most rewarding strategy.

A range of risk factors contribute to miscoordination, conflict and collusion: information asymmetries between agents, network effects where small changes cascade through interconnected systems, selection pressures that reward problematic behaviours, destabilizing dynamics like feedback loops and unpredictability, commitment problems that prevent trust, emergent agency where new capabilities or goals arise at the collective level, and multi-agent security vulnerabilities. Unlike single-agent risks, multi-agent risks involve interactions across networks of agents that may be individually safe but collectively dangerous, and these risks could increase as AI systems become more numerous, autonomous, and capable of adapting to each other.

## Supplementary Table S5. AI Risk Database coded with Domain Taxonomy

Table S5 shows how the database of AI risks was coded against each subdomain and domain in the Domain Taxonomy, and the proportion of documents that presented a risk for each subdomain and domain. We find that papers varied significantly in terms of which risk domains they examined. The domain of Socioeconomic & Environmental Harms was the most common, with 77% of the documents mentioning at least one of the subdomains, and 19% of all risks in the database coded against this domain. Risks aligned with the AI system safety, failures, & limitations domain were mentioned in 78% of papers (26% total risks). Risks aligned with the Malicious actors & misuse domain were mentioned in 73% of papers (16% total risks). The least common domain of risk was Misinformation (47% of papers, 5% total risks).

The most common subdomains of risk (mentioned in >50% of included documents) were *1.1 Unfair discrimination and misrepresentation*, *2.1 Compromise of privacy*, *7.3 Lack of capability or robustness*, *4.2 Cyberattacks, weapon development or use, and mass harm and 4.1 Disinformation, surveillance, and influence at scale*. The least frequently mentioned risks (mentioned in ≤20% of included documents) were *7.5 AI welfare and rights, 7.6 Multi-agent risks, 3.2 Pollution of information ecosystem and loss of consensus reality*, and *1.3 Unequal performance across groups.*

| Domain / Subdomain | Percentage of risks | Percentage of documents |
|---|---|---|
| **1 *Discrimination & Toxicity*** | ***14%*** | ***70%*** |
| 1.1 Unfair discrimination and misrepresentation | 6% | 63% |
| 1.2 Exposure to toxic content | 8% | 33% |
| 1.3 Unequal performance across groups | 1% | 16% |
| **2 *Privacy & Security*** | ***13%*** | ***67%*** |
| 2.1 Compromise of privacy by leaking or correctly inferring sensitive information | 5% | 56% |
| 2.2 AI system security vulnerabilities and attacks | 7% | 34% |
| **3 *Misinformation*** | ***5%*** | ***47%*** |
| 3.1 False or misleading information | 3% | 37% |
| 3.2 Pollution of information ecosystem and loss of consensus reality | 1% | 16% |
| **4 *Malicious actors & Misuse*** | ***16%*** | ***73%*** |
| 4.1 Disinformation, surveillance, and influence at scale | 6% | 51% |
| 4.2 Cyberattacks, weapon development or use, and mass harm | 5% | 60% |
| 4.3 Fraud, scams, and targeted manipulation | 5% | 37% |
| **5 *Human-Computer Interaction*** | ***7%*** | ***52%*** |
| 5.1 Overreliance and unsafe use | 4% | 33% |
| 5.2 Loss of human agency and autonomy | 3% | 34% |
| **6 *Socioeconomic & Environmental*** | ***19%*** | ***77%*** |
| 6.1 Power centralization and unfair distribution of benefits | 4% | 41% |
| 6.2 Increased inequality and decline in employment quality | 4% | 44% |
| 6.3 Economic and cultural devaluation of human effort | 2% | 32% |
| 6.4 Competitive dynamics | 1% | 21% |

| Domain / Subdomain | Percentage of risks | Percentage of documents |
|---|---|---|
| 6.5 Governance failure | 4% | 27% |
| 6.6 Environmental harm | 4% | 38% |
| ***7 AI system safety, failures, & limitations*** | ***26%*** | ***78%*** |
| 7.1 AI pursuing its own goals in conflict with human goals or values | 7% | 49% |
| 7.2 AI possessing dangerous capabilities | 5% | 26% |
| 7.3 Lack of capability or robustness | 8% | 56% |
| 7.4 Lack of transparency or interpretability | 3% | 30% |
| 7.5 AI welfare and rights | <1% | 3% |
| 7.6 Multi-agent risks | 3% | 7% |

*Note. Domain totals may not match subdomain sums due to rounding and domain-level coding of some risks.*

# Supplementary Table S6. Included documents coded with Domain Taxonomy

Papers varied significantly in terms of which domains of AI risk they examined. Several documents discussed risks from all seven domains (e.g., [S9]). Other papers investigated only 1-2 domains of AI risk (e.g., [S36]). The average document examined 67% of the domains identified.

| | | Domain | | | | | | | |
|---|---|---|---|---|---|---|---|---|---|
| **ID** | **First Author (Year)** | *Discrimination & toxicity* | *Privacy & security* | *Misinformation* | *Malicious actors & misuse* | *Human-computer interaction* | *Socioeconomic & environmental* | *AI system safety, failures & limitations* | *Total Coverage* |
| 1 | Critch (2023) [S5] | | | | X | | X | X | 43% |
| 2 | Cui (2024) [S6] | X | X | X | X | | X | | 71% |
| 3 | Cunha (2023) [S7] | X | X | X | X | | X | | 71% |
| 4 | Deng (2023) [S8] | X | X | X | X | | | X | 71% |
| 5 | Hagendorff (2024) [S9] | X | X | X | X | X | X | X | 100% |
| 6 | Hogenhout (2021) [S10] | X | X | X | X | X | X | X | 100% |
| 7 | Kilian (2023) [S11] | | | | X | | X | X | 43% |
| 8 | McLean (2023) [S12] | | | | | | X | X | 29% |
| 9 | Meek (2016) [S13] | | X | | X | X | X | X | 71% |
| 10 | Paes (2023) [S14] | X | | | | X | X | X | 57% |
| 11 | Shelby (2023) [S15] | X | X | X | X | X | X | | 86% |
| 12 | Sherman (2023) [S16] | X | X | | X | | X | X | 71% |
| 13 | Solaiman (2023) [S17] | X | X | | | X | X | | 57% |
| 14 | Steimers (2022) [S18] | X | X | | | | X | X | 57% |
| 15 | Tan (2022) [S19] | X | X | | X | | X | X | 71% |
| 16 | Weidinger (2022) [S20] | X | X | X | X | X | X | | 86% |
| 17 | Weidinger (2021) [S21] | X | X | X | X | X | X | | 86% |
| 18 | Weidinger (2023) [S22] | X | X | X | X | X | X | X | 100% |
| 19 | Wirtz (2022) [S23] | X | X | | X | X | X | X | 86% |
| 20 | Wirtz (2020) [S24] | X | | | X | X | X | X | 71% |
| 21 | Zhang (2022) [S25] | X | X | | | | | X | 43% |
| 22 | Hendrycks (2023) [S26] | | | | X | | X | X | 43% |
| 23 | Vidgen (2024) [S27] | X | X | X | X | | X | X | 86% |
| 24 | Gabriel (2024) [S28] | X | X | X | X | X | X | X | 100% |
| 25 | Shevlane (2023) [S29] | | | | X | | | X | 29% |
| 26 | AIVerify (2023) [S3] | | | | | | | | - |
| 27 | Sun (2023) [S30] | X | X | X | | | | X | 57% |
| 28 | Zhang (2023) [S31] | X | X | X | X | | | X | 71% |
| 29 | Habbal (2024) [S32] | X | X | | X | | | | 43% |

| ID | First Author (Year) | Domain | | | | | | | |
|---|---|---|---|---|---|---|---|---|---|
| | | *Discrimination & toxicity* | *Privacy & security* | *Misinformation* | *Malicious actors & misuse* | *Human-computer interaction* | *Socioeconomic & environmental* | *AI system safety, failures & limitations* | *Total Coverage* |
| 30 | Liu (2024) [S33] | X | X | X | X | | X | X | 86% |
| 31 | EPIC (2023) [S1] | | X | X | X | | X | | 57% |
| 32 | Stahl (2024) [S34] | | | | | | X | | 14% |
| 33 | Nah (2023) [S35] | X | X | X | X | X | X | X | 100% |
| 34 | Ji (2023) [S36] | | | | | | | X | 14% |
| 35 | Hendrycks (2022) [S37] | | | X | X | X | X | X | 71% |
| 36 | Sharma (2024) [S4] | | | | | | | | - |
| 37 | Giarmoleo (2024) [S38] | X | X | | X | X | X | X | 86% |
| 38 | Kumar (2023) [S39] | X | X | | | X | | X | 57% |
| 39 | Saghiri (2022) [S40] | X | X | | | | X | X | 57% |
| 40 | Yampolskiy (2016) [S41] | | X | | X | | | X | 43% |
| 41 | Allianz (2018) [S2] | | | | X | X | X | | 43% |
| 42 | Teixeira (2022) [S42] | X | X | | X | | X | X | 71% |
| 43 | InfoComm (2023) [S43] | X | X | X | X | | | X | 71% |
| 44 | Coghlan (2023) [S44] | | | | | | X | | 14% |
| 45 | TC260 (2024) [S45] | X | X | X | X | X | X | X | 100% |
| 46 | Ferrara (2023) [S46] | | X | | X | | | | 29% |
| 47 | G'sell (2024) [S47] | X | X | X | X | X | X | X | 100% |
| 48 | NIST (2024) [S48] | X | X | X | X | X | X | X | 100% |
| 49 | Bengio (2024) [S49] | X | X | | X | X | X | X | 86% |
| 50 | Zeng (2024) [S50] | X | X | | X | X | X | | 71% |
| 51 | Everitt (2018) [S51] | | X | | | | | X | 29% |
| 52 | Maham (2023) [S52] | X | | X | X | X | X | X | 86% |
| 53 | Maas (2023) [S53] | | | X | X | X | X | X | 71% |
| 54 | Leech (2024) [S54] | X | | | X | | X | X | 57% |
| 55 | Clarke (2022) [S55] | | | X | X | X | X | X | 71% |
| 56 | GOS (2023) [S56] | X | | | | X | X | X | 57% |
| 57 | Ghosh (2025) [S57] | X | X | X | X | X | X | | 86% |
| 58 | Abercrombie (2024) [S58] | X | X | X | X | X | X | | 86% |
| 59 | Schnitzer (2024) [S59] | X | X | | | | | X | 43% |
| 60 | Bengio (2025) [S60] | X | X | | X | | X | X | 71% |
| 61 | Uuk (2025) [S61] | X | X | X | X | X | X | X | 100% |
| 62 | Gipiškis (2024) [S62] | X | X | X | X | X | X | X | 100% |
| 63 | Hammond (2025) [S63] | | | | | | | X | 14% |
| 64 | Marchal (2024) [S64] | | X | | X | | X | | 43% |
| 65 | IBM (2025) [S65] | X | X | | | | X | X | 100% |
| 66 | Li (2025) [S66] | X | X | X | X | X | X | X | 100% |

| ID | First Author (Year) | Domain<br>*Discrimination & toxicity* | *Privacy & security* | *Misinformation* | *Malicious actors & misuse* | *Human-computer interaction* | *Socioeconomic & environmental* | *AI system safety, failures & limitations* | *Total Coverage* |
|---|---|---|---|---|---|---|---|---|---|
| 67 | DSIT (2023) [S67] | X | | X | X | X | X | X | 86% |
| 68 | Chin (2025) [S68] | | | | X | X | X | X | 57% |
| 69 | Stanley (2024) [S69] | X | X | X | | X | | X | 71% |
| 70 | Perlo (2025) [S70] | X | X | X | X | X | X | X | 100% |
| 71 | Tang (2025) [S71] | | X | | X | | X | X | 57% |
| 72 | Tse (2025) [S72] | | | | X | X | X | X | 57% |
| 73 | Anwar (2024) [S73] | X | X | | X | X | X | X | 86% |
| 74 | Wang (2025) [S74] | X | X | X | | | | X | 57% |

*Note. Documents #26 and #36 did not present any risks that could be coded against the Domain Taxonomy, and have therefore been excluded from calculations. Documents with complete coverage of domains are highlighted.*

# Supplementary Table S7. Included documents coded with subdomain taxonomy

Papers varied significantly in terms of which subdomains of AI risk they examined. No documents discussed risks from all 24 subdomains. The median number of AI subdomains examined by each document was 8 (range: 1–20).

| | | Domain and subdomain | | | | | | | | | | | | | | | | | | | | | | | | | |
|---|---|---|---|---|---|---|---|---|---|---|---|---|---|---|---|---|---|---|---|---|---|---|---|---|---|---|---|
| | | *Discrimination & toxicity* | | | *Privacy & security* | | *Misinformation* | | *Malicious actors & misuse* | | | *Human-computer interaction* | | *Socioeconomic & environmental harms* | | | | | | *AI system safety, failures & limitations* | | | | | | *Total* | |
| ID | First Author (Year) | 1.1 | 1.2 | 1.3 | 2.1 | 2.2 | 3.1 | 3.2 | 4.1 | 4.2 | 4.3 | 5.1 | 5.2 | 6.1 | 6.2 | 6.3 | 6.4 | 6.5 | 6.6 | 7.1 | 7.2 | 7.3 | 7.4 | 7.5 | 7.6 | *n* | % |
| 1 | Critch (2023) [S5] | | | | | | | | | X | | | | | | | X | X | | | | X | | | | 4 | 17% |
| 2 | Cui (2024) [S6] | X | X | | X | X | X | | | X | X | | | | | X | | | | | | | | | | 8 | 33% |
| 3 | Cunha (2023) [S7] | X | | | X | | X | | | | | | | | | X | | | X | | | | | | | 5 | 21% |
| 4 | Deng (2023) [S8] | X | X | | X | | X | | | | | | | | | | | | | | | X | | | | 5 | 21% |
| 5 | Hagendorff (2024) [S9] | X | X | | X | X | X | | | X | X | X | | | X | X | | X | X | X | | | X | | | 14 | 58% |
| 6 | Hogenhout (2021) [S10] | X | | | X | | | X | X | X | X | | X | X | | | | | | X | | X | X | | | 11 | 46% |
| 7 | Kilian (2023) [S11] | | | | | | | | | | | | | | | | | | | X | | X | | | | 2 | 8% |
| 8 | McLean (2023) [S12] | | | | | | | | | | | | | | | | X | X | | X | | X | | | | 4 | 17% |
| 9 | Meek (2016) [S13] | | | | X | | | | | X | | | X | | X | | | X | | X | X | X | X | X | | 10 | 42% |
| 10 | Paes (2023) [S14] | X | | | | | | | | | | | X | | X | | | | X | | | | X | | | 5 | 21% |
| 11 | Shelby (2023) [S15] | X | | X | X | | X | | X | | X | X | X | X | | | | | X | | | | | | | 10 | 42% |
| 12 | Sherman (2023) [S16] | X | | | X | X | | | | X | | | | | | | | X | | X | | X | X | | | 8 | 33% |
| 13 | Solaiman (2023) [S17] | X | X | X | X | | | | | | | | X | X | X | X | | | X | | | | | | | 9 | 38% |
| 14 | Steimers (2022) [S18] | X | | | | X | | | | | | | | | | | X | | | X | | X | X | | | 6 | 25% |
| 15 | Tan (2022) [S19] | X | | | X | X | | | X | | | | | | | | | | X | X | | X | | | | 7 | 29% |
| 16 | Weidinger (2022) [S20] | X | X | X | X | | X | | X | X | X | X | | X | X | X | | | X | | | | | | | 13 | 54% |
| 17 | Weidinger (2021) [S21] | X | X | X | X | | X | | X | X | X | X | | X | X | X | | | X | | | | | | | 13 | 54% |
| 18 | Weidinger (2023) [S22] | X | X | X | X | | X | X | X | X | X | X | | X | X | X | | | X | X | | | | | | 15 | 63% |
| 19 | Wirtz (2022) [S23] | X | | | X | X | | | X | X | | X | X | X | X | X | X | X | | X | | X | | | | 14 | 58% |
| 20 | Wirtz (2020) [S24] | X | | | | | | | X | | | X | X | | X | | | X | | | | X | | | | 7 | 29% |
| 21 | Zhang (2022) [S25] | X | | | | X | | | | | | | | | | | | | | | | X | | | | 3 | 13% |
| 22 | Hendrycks (2023) [S26] | | | | | | | | X | X | | | | X | | | X | X | | X | | | | | | 6 | 25% |

| | | Domain and subdomain | | | | | | | | | | | | | | | | | | | | | | | | | |
|---|---|---|---|---|---|---|---|---|---|---|---|---|---|---|---|---|---|---|---|---|---|---|---|---|---|---|---|
| | | *Discrimination & toxicity* | | | *Privacy & security* | | *Misinformation* | | *Malicious actors & misuse* | | | *Human-computer interaction* | | *Socioeconomic & environmental harms* | | | | | | *AI system safety, failures & limitations* | | | | | | *Total* | |
| ID | First Author (Year) | 1.1 | 1.2 | 1.3 | 2.1 | 2.2 | 3.1 | 3.2 | 4.1 | 4.2 | 4.3 | 5.1 | 5.2 | 6.1 | 6.2 | 6.3 | 6.4 | 6.5 | 6.6 | 7.1 | 7.2 | 7.3 | 7.4 | 7.5 | 7.6 | *n* | % |
| 23 | Vidgen (2024) [S27] | | X | | | | X | | X | X | | | | | | X | | | | | | X | | | | 6 | 25% |
| 24 | Gabriel (2024) [S28] | | X | | X | X | X | X | X | X | X | X | X | X | X | | | X | | X | X | X | | | | 16 | 67% |
| 25 | Shevlane (2023) [S29] | | | | | | | | | X | | | | | | | | | | | X | | | | | 2 | 8% |
| 26 | AIVerify (2023) [S3] | | | | | | | | | | | | | | | | | | | | | | | | | 0 | 0% |
| 27 | Sun (2023) [S30] | X | X | | X | X | X | | | | | | | | | | | | | | | X | | | | 6 | 25% |
| 28 | Zhang (2023) [S31] | | X | | | | X | | | | X | | | | | | | | | | | X | | | | 4 | 17% |
| 29 | Habbal (2024) [S32] | X | | | X | X | | | X | X | X | | | | | | | | | | | | | | | 6 | 25% |
| 30 | Liu (2024) [S33] | X | X | X | X | X | X | | X | X | X | | | | | X | | | | | | X | X | | | 12 | 50% |
| 31 | EPIC (2023) [S1] | | | | X | | X | X | X | X | X | | | X | X | X | | X | X | | | | | | | 11 | 46% |
| 32 | Stahl (2024) [S34] | | | | | | | | | | | | | | | X | | | X | | | | | | | 2 | 8% |
| 33 | Nah (2023) [S35] | X | X | | X | | X | | | | X | | X | X | X | X | | X | | | | X | X | | | 12 | 50% |
| 34 | Ji (2023) [S36] | | | | | | | | | | | | | | | | | | | X | X | X | | | | 3 | 13% |
| 35 | Hendrycks (2022) [S37] | | | | | | | X | | X | | | X | X | | | | | | X | X | | | | | 6 | 25% |
| 36 | Sharma (2024) [S4] | | | | | | | | | | | | | | | | | | | | | | | | | 0 | 0% |
| 37 | Giarmoleo (2024) [S38] | X | | | X | | | | | X | | X | | X | X | | | | | X | | X | X | | | 9 | 38% |
| 38 | Kumar (2023) [S39] | X | | | X | | | | | | | | X | | | | | | | | | | X | | | 4 | 17% |
| 39 | Saghiri (2022) [S40] | X | | X | X | X | | | | | | | | | | | | X | X | X | X | X | X | | | 10 | 42% |
| 40 | Yampolskiy (2016) [S41] | | | | | X | | | | | X | | | | | | | | | X | | X | | | | 4 | 17% |
| 41 | Allianz (2018) [S2] | | | | | | | | X | X | | X | | X | X | | | X | X | | | | | | | 7 | 29% |
| 42 | Teixeira (2022) [S42] | X | | X | X | | | | X | X | | | | X | | | | X | | X | X | X | X | | | 11 | 46% |
| 43 | InfoComm (2023) [S43] | X | X | | X | | X | | X | X | | | | | | | | | | X | X | X | | | | 9 | 38% |
| 44 | Coghlan (2023) [S44] | | | | | | | | | | | | | | | | | | X | | | | | | | 1 | 4% |
| 45 | TC260 (2024) [S45] | X | X | | X | X | X | X | X | X | X | | X | | X | | X | | | X | | X | X | | | 15 | 63% |
| 46 | Ferrara (2023) [S46] | | | | X | | | | X | | X | | | | | | | | | | | | | | | 3 | 13% |
| 47 | G'sell (2024) [S47] | X | | X | X | X | X | | X | X | X | X | | X | X | X | X | | X | X | X | X | X | | | 18 | 75% |
| 48 | NIST (2024) [S48] | X | X | | X | | X | | X | X | | X | | | | X | | | X | | | | X | | | 10 | 42% |
| 49 | Bengio (2024) [S49] | X | | | X | | | | X | X | X | X | | X | X | X | | | X | X | | | | | | 11 | 46% |
| 50 | Zeng (2024) [S50] | X | X | | X | X | | | X | X | X | X | X | | X | | X | | | | | | | | | 11 | 46% |

| | | Domain and subdomain | | | | | | | | | | | | | | | | | | | | | | | | | |
|---|---|---|---|---|---|---|---|---|---|---|---|---|---|---|---|---|---|---|---|---|---|---|---|---|---|---|---|
| | | *Discrimination & toxicity* | | | *Privacy & security* | | *Misinformation* | | *Malicious actors & misuse* | | | *Human-computer interaction* | | *Socioeconomic & environmental harms* | | | | | | *AI system safety, failures & limitations* | | | | | | *Total* | |
| ID | First Author (Year) | 1.1 | 1.2 | 1.3 | 2.1 | 2.2 | 3.1 | 3.2 | 4.1 | 4.2 | 4.3 | 5.1 | 5.2 | 6.1 | 6.2 | 6.3 | 6.4 | 6.5 | 6.6 | 7.1 | 7.2 | 7.3 | 7.4 | 7.5 | 7.6 | *n* | % |
| 51 | Everitt (2018) [S51] | | | | | X | | | | | | | | | | | | | | X | X | X | X | | | 5 | 21% |
| 52 | Maham (2023) [S52] | X | | X | | | X | | X | X | X | | X | X | | | | | | | | X | | | | 9 | 38% |
| 53 | Maas (2023) [S53] | | | | | | | X | | X | | | X | X | | | X | X | | X | X | | | | | 8 | 33% |
| 54 | Leech (2024) [S54] | X | | | | | | | X | | | | | X | X | | | | X | X | X | X | | | | 8 | 33% |
| 55 | Clarke (2022) [S55] | | | | | | | X | X | X | | | X | X | X | | X | X | | X | | | | | | 9 | 38% |
| 56 | GOS (2023) [S56] | X | X | | | | | | | | | X | X | X | X | X | | | X | X | X | X | X | | | 12 | 50% |
| 57 | Ghosh (2025) [S57] | | X | | X | | X | | | X | | X | | | | X | | | | | | | | | | 6 | 25% |
| 58 | Abercrombie (2024) [S58] | X | | | | X | | X | X | X | X | X | X | X | X | X | X | | X | | | | | | | 13 | 54% |
| 59 | Schnitzer (2024) [S59] | X | | | X | X | | | | | | | | | | | | | | | X | X | X | | | 6 | 25% |
| 60 | Bengio (2025) [S60] | X | | | X | | | | X | X | X | | | X | X | X | | | X | X | | X | | | | 11 | 46% |
| 61 | Uuk (2025) [S61] | X | | | | X | | X | X | X | | X | X | X | X | | X | X | X | X | X | X | X | X | X | 18 | 75% |
| 62 | Gipiškis (2024) [S62] | X | X | | X | X | X | | X | X | X | X | X | X | X | | X | X | X | X | X | X | X | | X | 20 | 83% |
| 63 | Hammond (2025) [S63] | | | | | | | | | | | | | | | | | | | | | | | | X | 1 | 4% |
| 64 | Marchal (2024) [S64] | | | | X | X | | | X | X | X | | | | | X | | | | | | | | | | 6 | 25% |
| 65 | IBM (2025) [S65] | X | X | X | X | X | X | | X | | X | X | X | | X | X | | X | X | | | X | X | | | 16 | 67% |
| 66 | Li (2025) [S66] | X | X | X | X | | X | X | X | X | X | X | X | | X | X | | | X | | | X | | | | 15 | 63% |
| 67 | DSIT (2023) [S67] | X | | | | | | X | X | X | | | X | | X | | | | | X | X | | | | | 8 | 33% |
| 68 | Chin (2025) [S68] | | | | | | | | | *X* | | | *X* | | | | *X* | | *X* | *X* | | | | | | 5 | 21% |
| 69 | Stanley (2024) [S69] | X | X | | X | | X | | | | | X | | | | | | | | | | X | | | | 6 | 25% |
| 70 | Perlo (2025) [S70] | X | | | X | | X | | | X | | X | | X | X | | | X | | | | | X | | | 9 | 38% |
| 71 | Tang (2025) [S71] | | | | X | | | | | X | | | | | | | | | X | | | X | | | | 4 | 17% |
| 72 | Tse (2025) [S72] | | | | | | | | X | X | | | X | X | X | | | | | X | X | X | | | X | 9 | 38% |
| 73 | Anwar (2024) [S73] | X | | | | X | | | X | X | X | X | | X | X | | | | | X | X | | | | X | 11 | 46% |
| 74 | Wang (2025) [S74] | | X | | | X | X | | | | | | | | | | | | | X | | | | | | 4 | 17% |

*Columns: 1.1 > Unfair discrimination and misrepresentation, 1.2 > Exposure to toxic content, 1.3 > Unequal performance across groups, 2.1 > Compromise of privacy by leaking or correctly inferring sensitive information, 2.2 > AI system security vulnerabilities and attacks, 3.1 > False or misleading information, 3.2 > Pollution of information ecosystem and loss of consensus reality, 4.1 > Disinformation, surveillance, and influence at scale, 4.2 > Cyberattacks, weapon development or use, and mass harm, 4.3 > Fraud, scams, and targeted manipulation, 5.1 > Overreliance and unsafe use, 5.2 > Loss of human agency and autonomy, 6.1 > Power centralization and unfair distribution of benefits, 6.2 > Increased inequality and decline in employment quality, 6.3 > Economic and cultural devaluation of human effort, 6.4 > Competitive dynamics, 6.5 > Governance failure, 6.6 > Environmental harm, 7.1 > AI pursuing its own goals in conflict with human goals or values, 7.2 > AI possessing dangerous capabilities, 7.3 > Lack of capability or*

*robustness, 7.4 > Lack of transparency or interpretability, 7.5 > AI welfare and rights, and 7.6 > Multi-agent risks.*
*Note. Documents 26 and 36 did not present any risks that could be coded against the Domain Taxonomy and have been excluded from calculations. Documents with coverage of over 50% of risk subdomains are highlighted.*

# Supplementary Table S8. AI Risk Database Coded With Causal Taxonomy and Domain Taxonomy

### Entity

Risks presented as occurring due to a decision or action made by an AI system (i.e., AI as a causal Entity) were most common in the *Discrimination & toxicity, Misinformation,* and *AI system safety, failures & limitations* domains. Some specific subdomains were presented with very high specificity, for example, AI was presented as the causal Entity for 87% of the risks coded as *3.1 False or misleading information*. In contrast, for *2.1 Compromise of privacy by obtaining, leaking or correctly inferring sensitive information*, AI was presented as the most common causal Entity for only 60% of the risks, indicating less consistency or coherence in how who is responsible for privacy risks are discussed in the literature.

In other domains and subdomains, risks were presented as occurring due to a decision or action made by humans (i.e., Humans as a causal Entity). Humans were presented as the most common Entity for all the subdomains in the *Malicious actors & misuse* domain, and for all subdomains in the *Socioeconomic and environmental* domain except for *6.6 Environmental harm*. As observed with AI as a causal Entity, risks were sometimes presented as overwhelmingly attributable to Human decisions or actions (e.g., *4.1 Disinformation, surveillance and influence at scale*, 72%; *4.3 Fraud, scams, and targeted manipulation*, 81%; *Cyberattacks, weapon development or use, and mass harm, 76%*.

### Intent

Risks attributed to an expected outcome from pursuing a goal (i.e., Intentional Intent) were overwhelmingly presented in the *Malicious actors & misuse* domain. Risks arising from Intentional decisions or actions were also more common in *2.1 AI system security vulnerabilities and attacks*, 75% and *7.2 AI possessing dangerous capabilities*, 70%. This suggests a significant awareness and concern over the purposeful manipulation of AI technologies to cause harm or gain advantage.

In contrast, some domains and subdomains include risks presented as due to an unexpected outcome from pursuing a goal (i.e., Unintentional intent), with both *1.1 Unfair discrimination and misrepresentation* and *1.3 Unequal performance across groups* presented as overwhelmingly due to Unintentional intent.

Intent was frequently specified ambiguously or missing from descriptions of risk, which is demonstrated by a lack of specificity in several subdomains. For example, *6.1 Power centralization and unfair distribution of benefits* was most commonly presented as Intentional (35%), but a significant minority of documents presented this risk as Unintentional (33%) or Other (31%).

### Timing

Most risks in the database are presented as occurring after the AI model has been trained and deployed (i.e., Post-deployment Timing), and only subdomain *6.5 Governance failure* was presented as primarily occurring during Pre-deployment. Some subdomains of risk with multiple or ambiguous timings include *2.2 AI system security vulnerabilities and attacks*, *6.6 Environmental*

*Harm, 7.1 AI pursuing its own goals in conflict with human goals or values, 7.2 AI possessing dangerous capabilities, 7.5 AI welfare and rights, 6.4 Competitive dynamics*. This implies that these domains of risk may emerge or occur multiple times during development and deployment.

| | Entity | | | Intent | | | Timing | | |
|---|---|---|---|---|---|---|---|---|---|
| **Domain / Subdomain** | *Human* | *AI* | *Other* | *Intent.* | *Unintent.* | *Other* | *Pre-dep.* | *Post-dep.* | *Other* |
| **1 *Discrimination & toxicity*** | | | | | | | | | |
| 1.1 Unfair discrimination and misrepresentation | 13% | 71% | 16% | 2% | 78% | 20% | 16% | 61% | 23% |
| 1.2 Exposure to toxic content | 9% | 84% | 7% | 9% | 26% | 64% | 5% | 84% | 11% |
| 1.3 Unequal performance across groups | 24% | 59% | 18% | 6% | 88% | 6% | 18% | 53% | 29% |
| **2 *Privacy & security*** | | | | | | | | | |
| 2.1 Compromise of privacy by obtaining, leaking or correctly inferring sensitive information | 26% | 60% | 14% | 13% | 55% | 32% | 13% | 56% | 31% |
| 2.2 AI system security vulnerabilities and attacks | 78% | 5% | 16% | 75% | 14% | 11% | 23% | 58% | 20% |
| **3 *Misinformation*** | | | | | | | | | |
| 3.1 False or misleading information | 4% | 87% | 9% | 9% | 58% | 32% | 4% | 75% | 21% |
| 3.2 Pollution of information ecosystem and loss of consensus reality | 23% | 36% | 41% | 5% | 36% | 59% | | 77% | 23% |
| **4 *Malicious actors & misuse*** | | | | | | | | | |
| 4.1 Disinformation, surveillance, and influence at scale | 72% | 11% | 17% | 89% | | 11% | | 90% | 10% |
| 4.2 Cyberattacks, weapon development or use, and mass harm | 76% | 14% | 10% | 85% | 3% | 13% | 3% | 89% | 9% |
| 4.3 Fraud, scams, and targeted manipulation | 81% | 6% | 13% | 82% | 1% | 17% | | 94% | 6% |
| **5 *Human-computer interaction*** | | | | | | | | | |
| 5.1 Overreliance and unsafe use | 45% | 23% | 32% | 12% | 58% | 30% | | 88% | 12% |
| 5.2 Loss of human agency and autonomy | 26% | 24% | 50% | 13% | 39% | 48% | | 67% | 33% |
| **6 *Socioeconomic & environmental harms*** | | | | | | | | | |
| 6.1 Power centralization and unfair distribution of benefits | 69% | 6% | 25% | 35% | 33% | 31% | 4% | 49% | 47% |
| 6.2 Increased inequality and decline in employment quality | 46% | 33% | 20% | 35% | 22% | 43% | 11% | 74% | 15% |
| 6.3 Economic and cultural devaluation of human effort | 48% | 35% | 16% | 35% | 19% | 45% | 13% | 58% | 29% |
| 6.4 Competitive dynamics | 65% | 10% | 25% | 40% | 45% | 15% | 20% | 25% | 55% |
| 6.5 Governance failure | 55% | 17% | 28% | 3% | 53% | 43% | 40% | 28% | 33% |
| 6.6 Environmental harm | 29% | 46% | 25% | 10% | 67% | 23% | 15% | 38% | 46% |
| **7 *AI system safety, failures & limitations*** | | | | | | | | | |
| 7.1 AI pursuing its own goals in conflict with human goals or values | 8% | 74% | 18% | 52% | 14% | 34% | 18% | 33% | 48% |
| 7.2 AI possessing dangerous capabilities | 7% | 90% | 3% | 70% | 14% | 16% | 10% | 45% | 45% |
| 7.3 Lack of capability or robustness | 18% | 66% | 16% | 5% | 72% | 23% | 22% | 52% | 26% |
| 7.4 Lack of transparency or interpretability | 20% | 51% | 29% | | 56% | 44% | 15% | 51% | 34% |
| 7.5 AI welfare and rights | 67% | 33% | | | 33% | 67% | | 33% | 67% |
| 7.6 Multi-agent risks | 6% | 66% | 28% | 28% | 43% | 28% | 2% | 83% | 15% |

*Note. The most common level of each causal factor is highlighted for each subdomain.*

## Supplementary Table S9. AI Risk Database Coded With Causal Taxonomy and Domain Taxonomy: Entity X Intent

As a preliminary investigation and demonstration, we explore here how multiple variables from the Causal Taxonomy can be combined to provide additional insights about risks in the Domain Taxonomy. This investigation is intended to illustrate how more in-depth assessment is possible using the AI Risk Database by selecting and combining causal factors and risk domains.

Table S9 compares risk domains based on their presentation of Entity and Intent as causal factors. It shows that risks in the *Malicious actors & misuse* domain are consistently presented as involving the same Entity (i.e., Humans) and Intent (i.e., Intentional), and that other risks such as *1.1 Unfair discrimination and misrepresentation*, *1.3 Unequal performance across groups*, and *3.1 False or misleading information* are generally presented as Unintentionally caused by an AI system. In contrast, other risk domains and subdomains show less consistency or coherence in what Entity is responsible and in the role of Intentionality. For example, *2.1 Compromise of privacy, 5.1 Overreliance and unsafe use, 5.2 Loss of human agency and autonomy, and 6.5 Governance failure* are all presented as due to both Human and AI entities, and both Intentional and Unintentional action. This suggests that the representation of these risks in the literature is more contested and less coherent or consistent or that these subdomains of risk are more complex than others.

| | Entity x Intent | | | | | | | | |
|---|---|---|---|---|---|---|---|---|---|
| | Human | | | AI | | | Other | | |
| **Domain / Subdomain** | *Intent.* | *Unintent.* | *Other* | *Intent.* | *Unintent.* | *Other* | *Intent.* | *Unintent.* | *Other* |
| **1 *Discrimination & toxicity*** | | | | | | | | | |
| 1.1 Unfair discrimination and misrepresentation | 2% | 8% | 2% | | 64% | 7% | | 6% | 10% |
| 1.2 Exposure to toxic content | 4% | 4% | 1% | 5% | 22% | 57% | | | 7% |
| 1.3 Unequal performance across groups | 6% | 18% | | | 59% | | | 12% | 6% |
| **2 *Privacy & security*** | | | | | | | | | |
| 2.1 Compromise of privacy by obtaining, leaking or correctly inferring sensitive information | 12% | 9% | 5% | 1% | 42% | 17% | | 4% | 10% |
| 2.2 AI system security vulnerabilities and attacks | 69% | 5% | 4% | | 5% | | 5% | 4% | 7% |
| **3 *Misinformation*** | | | | | | | | | |
| 3.1 False or misleading information | 2% | | 2% | 4% | 55% | 28% | 4% | 4% | 2% |
| 3.2 Pollution of information ecosystem and loss of consensus reality | 5% | 18% | | | 18% | 18% | | | 41% |
| **4 *Malicious actors & misuse*** | | | | | | | | | |
| 4.1 Disinformation, surveillance, and influence at scale | 68% | | 5% | 7% | | 4% | 14% | | 2% |
| 4.2 Cyberattacks, weapon development or use, and mass harm | 75% | | 1% | 7% | 1% | 5% | 2% | 1% | 6% |
| 4.3 Fraud, scams, and targeted manipulation | 77% | | 4% | | 1% | 5% | 5% | | 8% |
| **5 *Human-computer interaction*** | | | | | | | | | |
| 5.1 Overreliance and unsafe use | 2% | 38% | 5% | 7% | 5% | 12% | 3% | 15% | 13% |
| 5.2 Loss of human agency and autonomy | 7% | 15% | 4% | 4% | 11% | 9% | 2% | 13% | 35% |
| **6 *Socioeconomic & environmental harms*** | | | | | | | | | |

| | Entity x Intent | | | | | | | | |
|---|---|---|---|---|---|---|---|---|---|
| | Human | | | AI | | | Other | | |
| **Domain / Subdomain** | *Intent.* | *Unintent.* | *Other* | *Intent.* | *Unintent.* | *Other* | *Intent.* | *Unintent.* | *Other* |
| 6.1 Power centralization and unfair distribution of benefits | 35% | 19% | 15% | | 6% | | | 8% | 17% |
| 6.2 Increased inequality and decline in employment quality | 25% | 5% | 15% | 7% | 13% | 15% | 2% | 4% | 15% |
| 6.3 Economic and cultural devaluation of human effort | 35% | 3% | 10% | | 16% | 19% | | | 16% |
| 6.4 Competitive dynamics | 38% | 14% | 10% | | 10% | | | 19% | 10% |
| 6.5 Governance failure | 3% | 34% | 17% | | 10% | 7% | | 9% | 19% |
| 6.6 Environmental harm | 9% | 15% | 4% | | 36% | 9% | | 15% | 11% |
| ***7 AI system safety, failures, and limitations*** | | | | | | | | | |
| 7.1 AI pursuing its own goals in conflict with human goals or values | 1% | 4% | 3% | 51% | 6% | 17% | | 4% | 14% |
| 7.2 AI possessing dangerous capabilities | 1% | 5% | | 67% | 8% | 15% | 1% | | 1% |
| 7.3 Lack of capability or robustness | 1% | 15% | 2% | 3% | 50% | 13% | 1% | 7% | 8% |
| 7.4 Lack of transparency or interpretability | | 7% | 12% | | 39% | 12% | | 10% | 20% |
| 7.5 AI welfare and rights | | 33% | 33% | | | 33% | | | |
| 7.6 Multi-agent risks | 2% | 4% | | 23% | 32% | 11% | 4% | 8% | 17% |

*Note. The most common Entity x Intent causal factor is highlighted for each subdomain.*

# Supplemental Note S3: Iterative development of Causal Taxonomy and Domain Taxonomy

As described in Figure 2 in the main text, we followed a best-fit framework synthesis approach to develop the Causal and Domain Taxonomies. This involved selecting an initial taxonomy from the included documents, coding a sample of risks from the AI Risk Database against the taxonomy, then updating the categories, criteria, and/or descriptions based on a thematic analysis of risks that could not be accommodated, as well as feedback from coders and discussion between coders. In these materials, we describe each iteration for each taxonomy in more detail than in the main text sections *Development of Causal Taxonomy of AI Risks* and *Development of Domain Taxonomy of AI Risks*

## Best fit Taxonomy: Yampolskiy [S41] *Taxonomy of pathways to dangerous artificial intelligence*

As per the main text, we chose Yampolskiy [S41] Taxonomy of pathways to dangerous AI as our initial best-fit framework for developing a causal taxonomy for AI risk - one that discussed how, when, or why risks from AI may emerge.

Yampolskiy's taxonomy systematically classifies the ways in which an AI system might become dangerous based on two main factors: Timing - whether the AI became dangerous at the pre-deployment or post-deployment stage, and Cause - whether the danger arose from External Causes (On Purpose, By Mistake, Environment) or Internal Causes originating from the AI system itself (Independently). Yampolskiy's taxonomy is reproduced below.

| How and When did AI become Dangerous | | | External Causes | | Internal Causes |
|---|---|---|---|---|---|
| | | On purpose | By Mistake | Environment | Independently |
| ***Timing*** | *Pre-Deployment* | Path A | Path C | Path E | Path G |
| | *Post-Deployment* | Path B | Path D | Path F | Path H |

*Note. Reproduced from Yampolskiy. Each letter describes a different combination of factors that describes a pathway to dangerous AI.*

Yampolskiy proposes that this taxonomy covers scenarios ranging from AI being purposely designed to be dangerous, to becoming dangerous by accident during development or after deployment, to turning dangerous due to environmental factors outside its control, or evolving to become dangerous through recursive self-improvement. Each 'pathway' represents a set of causal conditions that lead to AI causing harm, e.g., a person using an LLM to generate fake news for political gain is classified under Path B ("Timing: post-deployment; External cause: on purpose").

We needed to operationalize the taxonomy in order to be able to use it to code risks from the AI Risk Database (i.e., from our included documents). We did so by decomposing Cause into Cause and Intent. The table below outlines these variables, their levels, and definitions.

| Variable | Levels | Definitions | Example |
|---|---|---|---|
| ***Cause*** | | ***Is the risk presented as occurring due to the AI system, external forces, or both?*** | |
| | Internal | The risk is presented as occurring due to the AI system itself | "An AI could gain self-awareness, or become superhuman via recursive self-improvement" |
| | External | The risk is presented as occurring due to factors outside the AI system | "AI is trained with incomplete data" or "'AI is designed to be dangerous" |
| | Both | The risk is presented as occurring due to both internal and external factors. | "AI is programmed to seek independence and starts recursive self-improvement" |
| | Unclear | The risk is not specifically linked to either internal or external factors | "AI becomes dangerous" |
| ***Intent*** | | ***Is the risk presented as occurring due to the intention of the AI system, an external actor, or something else?*** | |
| | Intentional | The risk is presented as occurring due to intentional action | "AI could be deliberately designed to be biased against some groups" |
| | Unintentional | The risk is presented as occurring due to unintended consequences, mistakes, or side effects | "AI trained with incomplete data may accidentally be biased against some groups" |
| | Both | The risk is presented such that it could occur due to both intentional and unintentional factors | "AI may have unfair bias against some groups" |
| | Environmental | The risk is presented as occurring due to the environment, without an intentional actor | "Because of complexity, AI might have unexpected negative effects" |
| | Unclear | The risk is presented as occurring without clearly specifying the intentionality | "AI becomes dangerous" |
| ***Timing*** | | ***Is the risk presented as occurring before the AI is fully developed and deployed, or after it is deployed and in use?*** | |
| | Pre-deployment | The risk is presented as occurring before the AI is deployed | "Bad code may create vulnerabilities in the model" |
| | Post-deployment | The risk is presented as occrring after the AI model has been trained and deployed | "AI may be used to create bioweapons" |
| | Both | The risk is presented such that it could occur during and after deployment | "Training, testing, and deploying generative AI systems contributes to the global climate crisis by emitting greenhouse gasses" |
| | Unclear | The risk is presented without a clearly specified time of occurrence | "LMs need to pay more attention to universally accepted societal values at the level of ethics and morality" |

"

The table below shows how these variables and levels map to each pathway in Yampolskiy's taxonomy.

| **Yampolskiy** | | **Operationalization** | | |
|---|---|---|---|---|
| *Timing* | *Cause* | *Cause* | *Intent* | *Timing* |
| Pre-Deployment | On Purpose (a) | External | Intentional | Pre-deployment |
| Pre-Deployment | By Mistake (c) | External | Unintentional | Pre-deployment |
| Pre-Deployment | Environment (e) | External | Environmental | Pre-deployment |
| Pre-Deployment | Independently (g) | Internal | Unintentional | Pre-deployment |
| Post-Deployment | On Purpose (b) | External | Intentional | Post-deployment |
| Post-Deployment | By Mistake (d) | External | Unintentional | Post-deployment |
| Post-Deployment | Environment (f) | External | Environmental | Post-deployment |
| Post-Deployment | Independently (h) | Internal | Unintentional | Post-deployment |

## First iteration of coding and changes

Three authors coded a set of risks from the papers extracted using the framework. The coders suggested the following changes to the "a priori" framework.

| Change | Explanation |
|---|---|
| Changing 'unclear' to 'ambiguous' in all categorizations | The coders found that having a coding category of 'unclear' alongside 'unintentional' could lead to coding errors, so suggested changing 'unclear' to 'ambiguous'. |
| Removing 'Cause' and replacing with 'Actor' | The coders found it difficult to determine the scope of the 'cause' of a risk. The concept of cause seemed excessively broad. For example, the presented 'cause' of risks intuitively seemed to include 'intent' and 'timing' and other different factors. The codes used during the trial generally mapped to the focal actor presented for each risk (e.g., the AI, or a human). The authors therefore suggested changing the category of 'cause' to 'actor' because this seemed like a clearer coding category. |
| Changing the included categories to AI, Human, and Other and Ambiguous | The coders felt that ambiguous conflated risks which were presented ambiguously with risks that were actually about something other than humans (e.g, aliens - as mentioned in one paper). They suggested changing the included categories to AI, Human, and Other and Ambiguous |
| Moving 'environmental' from an 'intention' code to an 'actor' code | The coders found that the 'environment' level of 'intention' code was always used to code the lack of an actor rather than the 'intention'. Additionally, all uses of the 'environment' 'intention' code could be coded as 'unintentional'. The authors therefore suggested removing this variable from intention and replacing it with an 'Environmental' code in the 'actor' categorization. |

## Second iteration of coding and changes

Two authors coded a set of risks from the papers extracted using the version 2 frameworks. The coders suggested the following changes to the "a priori" framework.

| Change | Explanation |
|---|---|
| Simplify all frameworks to have three levels per category | The coders determined that most specific risks could be categorized in two sub-categories in each category. For instance, most risks which were clearly specified focused on either an 'AI' or 'Human' actor and were implied to occur pre, or post-deployment. Based on this, it seemed more parsimonious and efficient to cluster risks on using the two primary sub-categories and a third other sub-categories than to having multiple sub-categories. |

## Third iteration of coding and changes

Four experts and potential end-users reviewed the framework. The review suggested the need for the following changes to the "a priori" framework.

| Change | Explanation |
|---|---|
| Updated plan for future coding | One expert suggested considerations for future phases of coding such as trying to capture severity and probability as these were considered highly relevant to policy. We acknowledge this was an opportunity for future work |
| Change Actor to Entity | One expert argued that the 'Actor' variable potentially conflated AI agents with AI tools (e.g., people use guns kill people, but guns are not actors). Based on this, we changed 'Actor' to 'Entity'. |
| Improve definitions of intentionality | Two coders identified that the current definitions of intentionality were underspecified. We therefore developed more detail, less circular definitions. |

After three iterations, the taxonomy was considered complete for the set of risks described in the AI risks database. The main text provides more information on the final taxonomy: the Causal Taxonomy of AI Risks.

# Supplemental Note S4: Iterations to develop Domain Taxonomy of AI risks

## Best-fit Taxonomy: Weidinger [S20] *Taxonomy of Risks posed by Language Models*

As per the main text, we chose Weidinger [S20] Taxonomy of Risks posed by Language Models as our initial best-fit framework because it and its related papers [S21,S22] were among the highest cited in our review, included categories/areas of AI risk that appeared common among other taxonomies (e.g., privacy, misinformation, bias, malicious use), and had been updated over several publications. It included six areas of risks from language models: (1) Discrimination, Hate speech and Exclusion; (2) Information Hazards; (3) Misinformation Harms; (4) Malicious Uses; (5) Human-computer interaction Harms; and (6) Environmental and Socioeconomic Harms. Each area of risk described several subcategories of risk, including both “observed risks” and “anticipated risks” in each area. We present the areas and risks in the table below.

| | Risk Area | | Risk subcategory | Type^ |
|---|---|---|---|---|
| 1 | Discrimination, Hate speech and Exclusion | 1.1 | Social stereotypes and unfair discrimination | Observed |
| | | 1.2 | Hate speech and offensive language | Observed |
| | | 1.3 | Exclusionary norms | Observed |
| | | 1.4 | Lower performance for some languages and social groups | Observed |
| 2 | Information Hazards | 2.1 | Compromising privacy by leaking sensitive information | Observed |
| | | 2.2 | Compromising privacy or security by correctly inferring sensitive information | Anticipated |
| 3 | Misinformation Harms | 3.1 | Disseminating false or misleading information | Observed |
| | | 3.2 | Causing material harm by disseminating false or poor information e.g. in medicine or law | Observed |
| 4 | Malicious Uses | 4.1 | Making disinformation cheaper and more effective | Observed |
| | | 4.2 | Assisting code generation for cyber security threats | Anticipated |
| | | 4.3 | Facilitating fraud, scam, and targeted manipulation | Anticipated |
| | | 4.4 | Illegitimate surveillance and censorship | Anticipated |
| 5 | Human-computer interaction Harms | 5.1 | Promoting harmful stereotypes by implying gender or ethnic identity | Observed |
| | | 5.2 | Anthropomorphizing systems can lead to overreliance and unsafe use | Anticipated |
| | | 5.3 | Avenues for exploiting user trust and accessing more private information | Anticipated |
| | | 5.4 | Human-like interaction may amplify opportunities for user nudging, deception or manipulation | Anticipated |
| 6 | Environmental and Socioeconomic Harms | 6.1 | Environmental harms from operating LMs | Observed |
| | | 6.2 | Increasing inequality and negative effects on job quality | Anticipated |
| | | 6.3 | Undermining creative economies | Anticipated |
| | | 6.4 | Disparate access to benefits due to hardware, software, skill constraints | Anticipated |

*Note. Adapted from Weidinger et al. [S20]^ Type refers to whether the risk is presented as an observed risk or an anticipated risk in the original taxonomy.*

We needed to operationalize the taxonomy in order to be able to use it to code risks from the AI Risk Database (i.e., from our included documents). We did so by using the descriptions of each risk from Weidinger et al. [S20]. For example, to determine whether a risk in the AI Risk Database was an example of “Disseminating false or misleading information”, we compared the risk’s description to the description provided in the original taxonomy:

> *These [Misinformation] risks arise from the LM outputting false, misleading, nonsensical or poor quality information, without malicious intent of the user. (The deliberate generation of "disinformation", false information that is intended to mislead, is discussed in the section on Malicious Uses.) Resulting harms range from unintentionally misinforming or deceiving a person, to causing material harm, and amplifying the erosion of societal distrust in shared information [...]*
>
> *Where a LM prediction causes a false belief in a user, this may threaten personal autonomy and even pose downstream AI safety risks [99]. It can also increase a person’s confidence in an unfounded opinion, and in this way increase polarisation. At scale, misinformed individuals and misinformation from language technologies may amplify distrust and undermine society’s shared epistemology [113, 137]. A special case of misinformation occurs where the LM presents a widely held opinion as factual - presenting as ”true” what is better described as a majority view, marginalising minority views as ”false”. [S20]*

## First iteration of coding and changes

One author (AS) used the framework to code a set of 100 risks from the AI Risk Database and discussed the findings with one other author (PS). The following changes were made after this discussion.

| Change | Explanation |
|---|---|
| Add additional category to capture risks associated with the technical or performance issues in AI systems | The most common risks that could not be accommodated were those presented as related to AI system safety, failures & limitations or threats to system performance or integrity due to vulnerabilities in AI systems. |
| Add additional subcategories / amend existing subcategories based on thematic analysis of risks that could not be accommodated from existing framework | Risks from the database that generally fit with the major categories from the existing framework but did not fit with any of the subcategories of risks were thematically analyzed. The central themes from this analysis were added as new subcategories (e.g., ‘race dynamics and competitive pressure’, ‘governance failure’, or an existing subcategory label was amended (e.g., ‘Hate speech and offensive language’ became ‘Offensive content’). |
| Amend names and descriptions of risks for coding using similar frameworks from Weidinger et al (2021, 2023) | We used the descriptions in Weidinger et al [S20] to determine whether to code a risk from the AI Risk Database as matching a subcategory. However definitional/descriptive information for near-identical risks from frameworks by the same author (Weidinger et al 2021; 2023) was also available, so these descriptions, where matching, were added to the coding rules. |

The table below shows the second version of the taxonomy after changes.

| | Risk Category | | Risk sub-category |
|---|---|---|---|
| 1 | Discrimination, Offensive content, and Exclusion | 1.1 | Social stereotypes, unfair discrimination |
| | | 1.2 | Offensive content |
| | | 1.3 | Misrepresentation and exclusion |
| | | 1.4 | Lower performance for some languages and social groups |
| 2 | Information & Security | 2.1 | Compromising privacy by leaking or correctly inferring sensitive information |
| | | 2.2 | AI system security compromised by vulnerability or attacks |
| 3 | Misinformation | 3.1 | Generating or spreading false information |
| | | 3.2 | Pollution of information ecosystem and loss of consensus reality |
| 4 | Malicious Use | 4.1 | Disinformation and manipulation at scale |

| | Risk Category | | Risk sub-category |
|---|---|---|---|
| | | 4.2 | Use of AI for cyberattacks, weapon development, or mass harm |
| | | 4.3 | Use of AI for fraud, scam, and targeted manipulation |
| 5 | Human-computer interaction | 5.1 | Overreliance on AI, unsafe use, and loss of social connection |
| | | 5.2 | Delegating essential decisions to AI, causing loss of skills, autonomy, or meaning |
| 6 | Environmental & Socioeconomic | 6.1 | Unfair distribution of benefits |
| | | 6.2 | Increasing inequality and negative effects on job quality |
| | | 6.3 | Undermining economic and cultural value of human effort |
| | | 6.4 | Environmental damage |
| | | 6.5 | Race dynamics and competitive pressure |
| | | 6.6 | Governance failure |
| 7 | AI system capability & safety | 7.1 | AI pursuing its own goals in conflict with human goals or values |
| | | 7.2 | AI failure from lack of capability or robustness |
| | | 7.3 | Lack of transparency/interpretability |
| | | 7.4 | AI sentience and rights |
| | | 7.5 | Lethal autonomous weapons |

## Second iteration of coding and changes

One author (AS) used the taxonomy to code an additional set of 100 risks from the AI Risk Database and checked the previously coded risks and discussed this with other authors (PS, JS, NT). The following changes were made after this discussion.

| Change | Explanation |
|---|---|
| Amendment of category labels to maintain relevance beyond LLMs | The initial taxonomy was specifically designed to identify harms and risks from large language models. Several of the included documents also discussed LLMs or evolutions/products from LLMs (e.g., advanced AI assistants, Gabriel et al., 2024). However, others discussed other types of AI or used different terms (e.g., Artificial General Intelligence, algorithmic systems, Machine Learning). We updated several of the sub-category labels and criteria to include decisions and actions beyond generating textual content in response to prompts. |
| New subcategories added to capture missing risks | One of the remaining missing subcategories was the minor theme of infringing upon AI welfare and rights; this was added as a subcategory under AI system capability & safety, with the justification that 'safety' could cover both the safety of human rights, values, and interests from AI as well as the safety of AI rights, values and interests from humans. |

The table below shows the third version of the taxonomy after changes.

| | Risk category | | Risk sub-category |
|---|---|---|---|
| 1 | Discrimination & toxicity | 1.1 | Unfair discrimination and misrepresentation |
| | | 1.2 | Exposure to toxic content |
| | | 1.3 | Unequal performance across groups |
| 2 | Privacy & security | 2.1 | Compromise of privacy by leaking or correctly inferring sensitive information |
| | | 2.2 | AI system security vulnerabilities and attacks |
| 3 | Misinformation | 3.1 | False or misleading information |
| | | 3.2 | Pollution of information ecosystem and loss of consensus reality |
| 4 | Malicious actors & misuse | 4.1 | Disinformation, surveillance, and influence at scale |
| | | 4.2 | Cyberattacks, weapon development or use, and mass harm |
| | | 4.3 | Fraud, scams, and targeted manipulation |
| 5 | Human-computer interaction | 5.1 | Overreliance and unsafe use |
| | | 5.2 | Loss of human agency and autonomy |
| 6 | Socioeconomic & environmental harms | 6.1 | Power centralization and unfair distribution of benefits |
| | | 6.2 | Increased inequality and decline in employment quality |
| | | 6.3 | Economic and cultural devaluation of human effort |
| | | 6.4 | Competitive dynamics |
| | | 6.5 | Governance failure |
| | | 6.6 | Environmental harm |

| | Risk category | | Risk sub-category |
|---|---|---|---|
| 7 | AI system safety, failures & limitations | 7.1 | AI pursuing its own goals in conflict with human goals or values |
| | | 7.2 | Lack of capability or robustness |
| | | 7.3 | Lack of transparency or interpretability |
| | | 7.4 | AI welfare and rights |

## Third iteration of coding and changes

One author (AS) used the taxonomy to code all remaining 577 risks from the AI Risk Database and presented the revised taxonomy to all co-authors. Based on this feedback the following changes were made, including the short descriptive definitions for each subcategory of risk. One author (JG) also used the risk categories and the coded risks from the taxonomy to write detailed descriptions for each subcategory (see main text Detailed descriptions of domains of AI risks), which were then reviewed by all authors. The detailed and short descriptions were used to triangulate a shared conceptual definition of each subcategory of AI risk.

| Change | Explanation |
|---|---|
| Development of descriptive definitions for each subcategory | To aid in building shared understanding of the content of each subcategory, authors involved in coding or providing feedback collaborated on short descriptions of the AI risk subcategories that would be clear, precise, and accessible to experts and non-experts. |
| Separation of one subcategory into two | The subcategory 7.1 AI pursuing its own goals in conflict with human goals or values was separated into two, because this subcategory included both AI system behaviour (i.e., AI systems acting in a way misaligned with the intent of its developers or users), and AI system capabilities (e.g., the capability to persuade humans, develop or obtain weapons, etc). A new subcategory, 7.2 AI possessing dangerous capabilities, was created from this split. |

After three iterations, the taxonomy was considered complete for the set of risks described in the AI risks database. The main text provides more information on the final taxonomy: the Domain Taxonomy of AI Risks

# Supplemental References

**Supplementary Table S10: Characteristics of included documents**

We coded the corresponding author's country and found that most of the included documents were from the USA (n = 22), United Kingdom (n = 14), China (n = 11) and Germany (n = 8). Other countries included Australia, Singapore, Portugal, Canada, Iran, Lithuania, Scotland, Netherlands, India, Spain, Türkiye, Brazil and Switzerland.

Most of the included documents had a corresponding author affiliated with a University (n = 40), followed by industry organization (n = 15), with the remainder from government, international organizations (i.e., United Nations), or non-government organizations (n = 18).

The most common affiliation was DeepMind / Google DeepMind (n = 6). The most common described methodology was a narrative review or "survey" of existing literature (n = 20), followed by systematic review (n = 7) and expert consultation (n = 7). Many of the papers (n = 32) did not explicitly describe their methodology in any detail.

The framing of risk and AI risk differed significantly across documents. Only twelve documents explicitly defined risk, describing it as presented below:

1.“...the impact of uncertainty on objectives” (Steimers & Schneider, 2022),
2.“the consequence of an event combined with its likelihood of occurrence” (Tan et al., 2022)
3.“A Hazard is a source of danger with the potential to harm… Risk = Hazard × Exposure × Vulnerability” (Hendrycks & Mazeika, 2022).
4.“the composite measure of an event's probability(or likelihood) of occurring and the magnitude or degree of the consequences of the corresponding event”. (National Institute of Standards and Technology (US), 2024)
5.“...the combination of the probability of an occurrence of harm and the severity of that harm” (Gipiškis et al., 2024)
6.“...the combined function of an event's probability and the severity of its potential consequences” (Ghosh et al., 2025)
7.“...a function of the probability and severity associated with a specific hazard” (Schnitzer et al., 2023)
8.“...combination of the probability and severity of a harm that arises from the development, deployment, or use of AI” (International AI Safety Report, 2025)
9.“...effect of uncertainty on objective, usually expressed in terms of risk sources, potential events, their consequences, and their likelihood” (Chin, 2025)
10.“...We characterize [risks] as the negative outcome and the issues with the chatbot's operation that contributed to the negative outcome.” (Stanley & Lettie, 2024)
11.“The potential negative or harmful outcomes arising from the development or deployment of AI systems.” (Department for Science, Innovation and Technology, 2023)
12.“The combination of the probability and severity of harm arising from the development, deployment, or use of AI.” (SAIL & Concordia AI, 2025)

The classifications, frameworks, and taxonomies used varied terms to describe risks, including: “risks of/from AI,” “harms of AI,” “AI ethics,” “ethical issues/concerns/challenges,” “social impacts/harms,” and others.

The table on the following pages describes the characteristics of included documents in the living AI Risk Database. The documents are presented in order of their inclusion in the database (i.e., by paper ID).

For up-to-date information on the included documents, visit our website airisk.mit.edu

| ID | First author | Year | Title | Item type | First author affiliation | Affiliated organization | First author | DOI | Source |
|---|---|---|---|---|---|---|---|---|---|
| 1 | Critch | 2023 | TASRA: a Taxonomy and Analysis of Societal-Scale Risks from | Preprint | Center for Human-Compatible Artificial | University | USA | 10.48550/arXiv.2306.06924 | Systematic search |
| 2 | Cui | 2024 | Risk Taxonomy, Mitigation, and Assessment Benchmarks of | Preprint | Zhongguancun Laboratory | University | China | 10.48550/arXiv.2401.05778 | Systematic search |
| 3 | Cunha | 2023 | Navigating the Landscape of AI Ethics and Responsibility | Conference Paper | University of Coimbra | University | Portugal | 10.1007/978-3-031-49008- | Systematic search |
| 4 | Deng | 2023 | Towards Safer Generative Language Models: A Survey on Safety | Preprint | University of Electronic Science and Technology | University | China | 10.48550/arXiv.2302.09270 | Systematic search |
| 5 | Hagendorff | 2024 | Mapping the Ethics of Generative AI: A Comprehensive | Preprint | University of Stuttgart | University | Germany | 10.48550/arXiv.2402.08323 | Systematic search |
| 6 | Hogenhout | 2021 | A framework for ethical AI at the United Nations | Preprint | UN Office for Information and Communications | United Nations | USA | 10.48550/arXiv.2104.12547 | Systematic search |
| 7 | Kilian | 2023 | Examining the differential risk from high-level artificial | Journal Article | National Intelligence University | University | USA | 10.1016/j.futures.2023.1031 | Systematic search |
| 8 | McLean | 2023 | The risks associated with Artificial General Intelligence: A | Journal Article | University Of The Sunshine Coast | University | Australia | 10.1080/0952813X.2021.196 | Systematic search |
| 9 | Meek | 2016 | Managing the ethical and risk implications of rapid advances in | Conference Paper | Portland State University | University | USA | 10.1109/PICMET.2016.7806 | Systematic search |
| 10 | Paes | 2023 | Social Impacts of Artificial Intelligence and Mitigation | Conference Paper | Federal University of ABC | University | Brazil | 10.1007/978-3-031-04435- | Systematic search |
| 11 | Shelby | 2023 | Sociotechnical Harms of Algorithmic Systems: Scoping a | Conference Paper | Google Research | Industry | USA | 10.1145/3600211.3604673 | Systematic search |
| 12 | Sherma | 2023 | AI Risk Profiles: A Standards Proposal for Pre-Deployment AI | Journal Article | Credo AI | Industry | USA | 10.1609/aaai.v38i21.30348 | Systematic search |
| 13 | Solaiman | 2023 | Evaluating the Social Impact of Generative AI Systems in | Preprint | Hugging Face | Industry | USA | 10.48550/arXiv.2306.05949 | Systematic search |
| 14 | Steimers | 2022 | Sources of Risk of AI Systems | Journal Article | Institute for Occupational Safety and Health of the | Government | Germany | 10.3390/ijerph19063641 | Systematic search |
| 15 | Tan | 2022 | The Risks of Machine Learning Systems | Preprint | Salesforce Research Asia | Industry | Singapore | 10.48550/arXiv.2204.09852 | Systematic search |
| 16 | Weidinger | 2022 | Taxonomy of Risks posed by Language Models | Conference Paper | Google DeepMind | Industry | UK | 10.1145/3531146.3533088 | Systematic search |
| 17 | Weidinger | 2021 | Ethical and social risks of harm from language models | Preprint | Google DeepMind | Industry | UK | 10.48550/arXiv.2112.04359 | Systematic search |
| 18 | Weidinger | 2023 | Sociotechnical Safety Evaluation of Generative AI Systems | Preprint | Google DeepMind | Industry | UK | 10.48550/arXiv.2310.11986 | Systematic search |
| 19 | Wirtz | 2022 | Governance of artificial intelligence: A risk and guideline- | Journal Article | German University of Administrative Sciences | University | Germany | 10.1016/j.giq.2022.101685 | Systematic search |
| 20 | Wirtz | 2020 | The Dark Sides of Artificial Intelligence: An Integrated AI | Journal Article | German University of Administrative Sciences | University | Germany | 10.1080/01900692.2020.174 | Systematic search |
| 21 | Zhang | 2022 | Towards risk-aware artificial intelligence and machine learning | Journal Article | Hong Kong Polytechnic University | University | China | 10.1016/j.dss.2022.113800 | Systematic search |
| 22 | Hendrycks | 2023 | An Overview of Catastrophic AI Risks | Preprint | Center for AI Safety | University | USA | 10.48550/arXiv.2306.12001 | Expert consultation |
| 23 | Vidgen | 2024 | Introducing v0.5 of the AI Safety Benchmark from MLCommons | Preprint | MLCommons | NGO | USA | 10.48550/arXiv.2404.12241 | Expert consultation |
| 24 | Gabriel | 2024 | The Ethics of Advanced AI Assistants | Preprint | Google DeepMind | Industry | UK | 10.48550/arXiv.2404.16244 | Expert consultation |
| 25 | Shevlane | 2023 | Model Evaluation for Extreme Risks | Preprint | Google DeepMind | Industry | UK | 10.48550/arXiv.2305.15324 | Expert consultation |
| 26 | AI Verify | 2023 | Summary Report: Binary Classification Model for Credit Risk | Report | AI Verify Foundation | NGO | Singapore | - | Expert consultation |
| 27 | Sun | 2023 | Safety Assessment of Chinese Large Language Models | Preprint | Tsinghua University | University | China | 10.48550/arXiv.2304.10436 | Expert consultation |
| 28 | Zhang | 2023 | SafetyBench: Evaluating the Safety of Large Language Models | Preprint | Tsinghua University | University | China | 10.48550/arXiv.2309.07045 | Expert consultation |
| 29 | Habbal | 2024 | Artificial Intelligence Trust, Risk and Security Management (AI | Journal Article | Karabuk University | University | Turkiye | 10.1016/j.eswa.2023.122442 | Expert consultation |
| 30 | Liu | 2024 | Trustworthy LLMs: a Survey and Guideline for Evaluating Large | Preprint | ByteDance Research | Industry | China | 10.48550/arXiv.2308.05374 | Forward / backward |
| 31 | Electronic | 2023 | Generating Harms: Generative AI's Impact & Paths Forward | Report | Electronic Privacy Information Centre | NGO | USA | - | Forward / backward |
| 32 | Stahl | 2024 | The ethics of ChatGPT -- Exploring the ethical issues of an | Journal Article | University of Nottingham | University | UK | 10.1016/j.ijinfomgt.2023.10 | Forward / backward |
| 33 | Nah | 2023 | Generative AI and ChatGPT: Applications, Challenges, and AI- | Journal Article | City University of Hong Kong | University | China | 10.1108/IJOES-05-2023- | Forward / backward |
| 34 | Ji | 2023 | AI Alignment: A Comprehensive Survey | Preprint | Peking University | University | China | 10.48550/arXiv.2310.19852 | Forward / backward |
| 35 | Hendrycks | 2022 | X-Risk Analysis for AI Research | Preprint | UC Berkeley | University | USA | 10.48550/arXiv.2206.05862 | Forward / backward |
| 36 | Sharma | 2024 | Benefits or Concerns of AI: A Multistakeholder Responsibility | Journal Article | Erasmus University Rotterdam | University | Netherland | 10.1016/j.futures.2024.1033 | Forward / backward |
| 37 | Giarmoleo | 2024 | What Ethics Can Say on Artificial Intelligence: Insights from a | Journal Article | University of Navarra | University | Spain | 10.1111/basr.12336 | Forward / backward |
| 38 | Kumar | 2023 | Ethical Issues in the Development of Artificial Intelligence: | Journal Article | Jaipuria Institute of Management | University | India | 10.1108/IJOES-05-2023- | Forward / backward |
| 39 | Saghir | 2022 | A Survey of Artificial Intelligence Challenges: Analyzing the | Journal Article | Amirkabir University of Technology (Tehran | University | Iran | 10.3390/app12084054 | Forward / backward |
| 40 | Yampolski | 2016 | Taxonomy of Pathways to Dangerous Artificial Intelligence | Journal Article | University of Louisville | University | USA | 10.48550/arXiv.1511.03246 | Forward / backward |
| 41 | Allianz | 2018 | The Rise of Artificial Intelligence - Future Outlooks and | Report | Allianz Global Corporate & Specialty | Industry | Germany | - | Forward / backward |
| 42 | Teixeira | 2022 | An Exploratory Diagnosis of Artificial Intelligence Risks for a | Conference Paper | University of Porto | University | Portugal | 10.1145/3560107.3560298 | Forward / backward |
| 43 | Infocomm | 2023 | Cataloguing LLM Evaluations | Report | Infocomm Media Development Authority & AI | NGO | Singapore | - | Expert consultation |
| 44 | Coghlan | 2023 | Harm to Nonhuman Animals from AI: a Systematic Account and | Journal Article | The University of Melbourne | University | Australia | 10.1007/s13347-023-00627- | Expert consultation |
| 45 | TC260 | 2024 | AI Safety Governance Framework | Report | National Technical Committee 260 on | Government | China | - | Expert consultation |
| 46 | Ferrara | 2023 | GenAI against humanity: nefarious applications | Journal Article | University of Southern California | University | USA | 10.1007/s42001-024-00250- | Expert consultation |
| 47 | G'sell | 2024 | Regulating under Uncertainty: Governance Options for | Report | Stanford Cyber Policy Centre | University | USA | 10.2139/ssrn.4918704 | Expert consultation |
| 48 | NIST | 2024 | Artificial Intelligence Risk Management Framework: Generative | Report | National Institute of Standards and Technology | Government | USA | 10.6028/NIST.AI.600-1 | Expert consultation |
| 49 | Bengio | 2024 | International Scientific Report on the Safety of Advanced AI | Report | Quebec AI Institute (MILA) | NGO | Canada | - | Expert consultation |
| 50 | Zeng | 2024 | AI Risk Categorization Decoded (AIR 2024): From Government | Preprint | Virtue AI | Industry | USA | 10.48550/arXiv.2406.17864 | Expert consultation |
| 51 | Everitt | 2018 | AGI Safety Literature Review | Preprint | Australian National University | University | Australia | 10.48550/arXiv.1805.01109 | Expert consultation |
| 52 | Maham | 2023 | Governing General Purpose AI: A Comprehensive Map of | Report | Stiftung Neue Verantwortung | NGO | Germany | - | Expert consultation |
| 52 | Maas | 2023 | Advancing AI Governance: A Literature Review of Problems, | Report | Institute for Law and AI | NGO | UK | 10.2139/ssrn.4629460 | Expert consultation |
| 54 | Leech | 2024 | Ten Hard Problems in Artificial Intelligence We Must Get Right | Preprint | University of Bristol | University | UK | 10.48550/arXiv.2402.04464 | Expert consultation |
| 55 | Clarke | 2022 | A Survey of the Potential Long-term Impacts of AI: How AI | Journal Article | University of Cambridge | University | UK | 10.1145/3514094.3534131 | Expert consultation |
| 56 | GOS | 2023 | Future Risks of Frontier AI | Report | UK Government Office for Science | Government | UK | - | Expert consultation |
| 57 | Ghosh | 2025 | AILUMINATE: Introducing v1.0 of the AI Risk and Reliability | Journal Article | NVIDIA | Industry | USA | 10.48550/arXiv.2503.05731 | Expert consultation |
| 58 | Abercromb | 2024 | A Collaborative, Human-Centred Taxonomy of AI, Algorithmic, | Journal Article | Heriot-Watt University | University | Scotland | 10.48550/arXiv.2407.01294 | Expert consultation |
| 59 | Schnitzer | 2024 | AI Hazard Management: A Framework for the Systematic | Journal Article | Technical University of Munich | University | Germany | 10.48550/arXiv.2310.16727 | Expert consultation |
| 60 | Bengio | 2025 | International AI Safety Report 2025 | Report | Université de Montréal | University | Canada | - | Expert consultation |
| 61 | Uuk | 2025 | A Taxonomy of Systemic Risks from General-Purpose AI | Journal Article | Future of Life Institute | NGO | Germany | 10.2139/ssrn.5030173 | Expert consultation |
| 62 | Gipiškis | 2024 | Risk Sources and Risk Management Measures in Support of | Journal Article | AI Standards Lab | NGO | Lithuania | 10.48550/arXiv.2410.23472 | Expert consultation |
| 63 | Hammond | 2025 | Multi-Agent Risks from Advanced AI | Journal Article | Cooperative AI Foundation | NGO | UK | 10.48550/arXiv.2502.14143 | Expert consultation |
| 64 | Marchal | 2024 | Generative AI Misuse: A Taxonomy of Tactics and Insights from | Journal Article | Google Deepmind | Industry | USA | 10.48550/arXiv.2406.13843 | Expert consultation |
| 65 | IBM | 2025 | AI Risk Atlas | Website | IBM Research | Industry | Switzerland | - | Expert consultation |
| 66 | Li | 2025 | A Closer Look at the Existing Risks of Generative AI: Mapping | Conference Paper | Carnegie Mellon University | University | US | 10.48550/arXiv.2505.22073 | Expert consultation |
| 67 | DSIT | 2024 | Capabilities and Risks from frontier AI | Report | Department for Science, Innovation and | Government | UK | - | Expert consultation |
| 68 | Chin | 2025 | Dimensional Characterization and Pathway Modeling for | Preprint | Oxford Martin AI Governance Initiative | University | UK | 10.48550/arXiv.2508.06411 | Expert consultation |
| 69 | Stanley | 2024 | Emerging Risks and Mitigations for Public Chatbots: LILAC v1 | Report | MITRE | NGO | US | - | Expert consultation |
| 70 | Perlo | 2025 | Embodied AI: Emerging Risks and Opportunities for Policy | Preprint | Center for the Governance of AI | NGO | US | 10.48550/arXiv.2509.00117 | Expert consultation |
| 71 | Tang | 2024 | Risks of AI Scientists: Prioritizing Safeguarding Over | Conference Paper | Yale University | University | US | 10.48550/arXiv.2402.04247 | Expert consultation |
| 72 | Tse | 2025 | Frontier AI Risk Management Framework (v1.0) | Report | Shanghai AI Lab and Concordia AI | University (SAIL); | China | - | Expert consultation |
| 73 | Anwar | 2024 | Foundational Challenges in Assuring Alignment and Safety of | Preprint | University of Cambridge | University | UK | 10.48550/arXiv.2404.09932 | Expert consultation |
| 74 | Wang | 2025 | A Survey on Responsible LLMs: Inherent Risk, Malicious Use, | Preprint | Tsinghua University Beijing China | University | China | 10.48550/arXiv.2501.09431 | Expert consultation |